\documentclass[11pt]{article}

\usepackage[final]{acl}

\usepackage{times}
\usepackage{latexsym}

\usepackage[T1]{fontenc}

\usepackage[utf8]{inputenc}

\usepackage{microtype}

\usepackage{inconsolata}

\usepackage{graphicx}

\usepackage{xspace}
\usepackage{colortbl}
\usepackage{multirow}
\usepackage{arydshln}
\usepackage{booktabs}
\usepackage{ulem}
\usepackage[inline]{enumitem}
\usepackage{scalerel}
\usepackage{tablefootnote}
\usepackage{siunitx}
\usepackage{tipa}
\usepackage{bbding}
\usepackage{cleveref}
\usepackage{float}
\usepackage[weather]{ifsym}
\usepackage{pgfplots}
\usepgfplotslibrary{fillbetween}
\pgfplotsset{compat=1.18}
\usepackage{tikz}
\usetikzlibrary{positioning, shapes, arrows.meta,fit,matrix}

\newcommand{\mphubert}{\mbox{\textsc{MauBERT}}\xspace}
\newcommand{\feat}{\mbox{\textsc{feat}}\xspace}
\newcommand{\phone}{\mbox{\textsc{phone}}\xspace}
\def\Fire{\scalerel*{\includegraphics{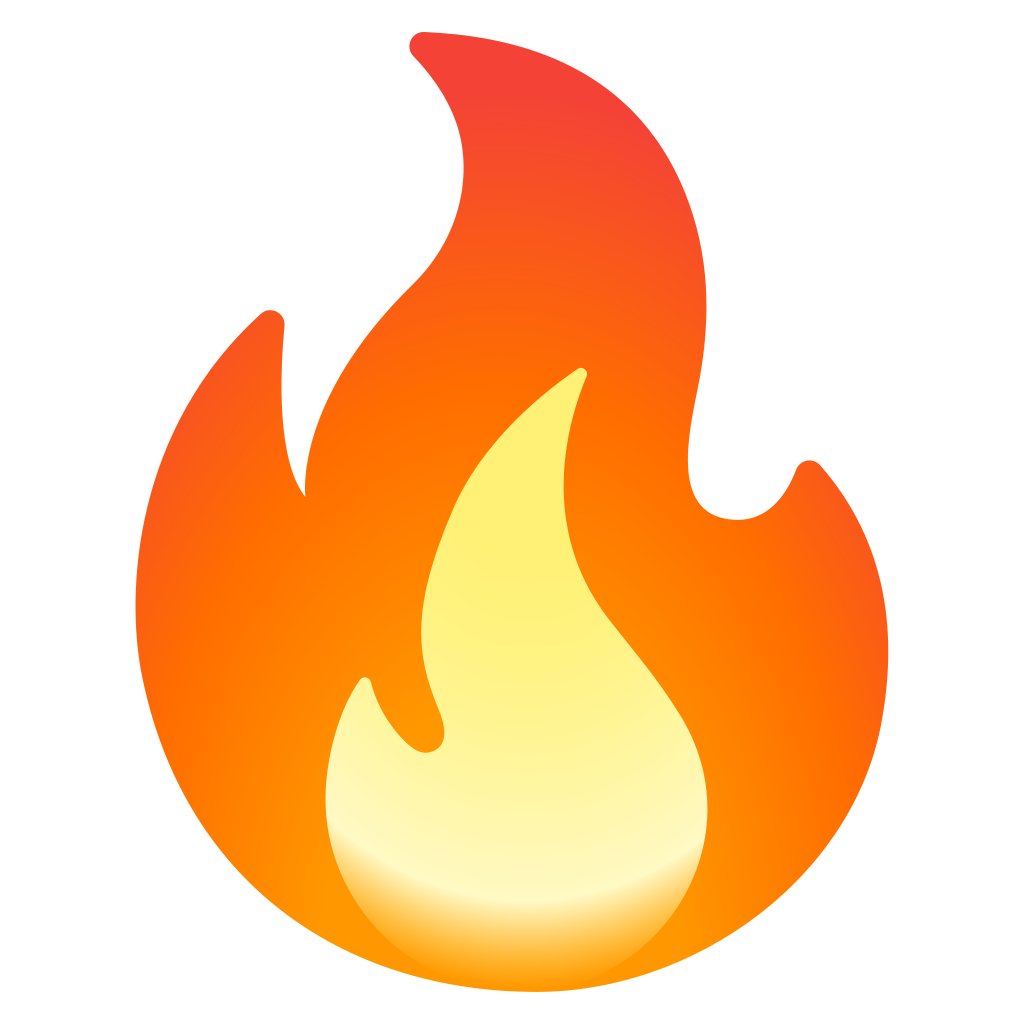}}{X}}

\makeatletter
\def\thanks#1{\protected@xdef\@thanks{\@thanks
    \protect\footnotetext{#1}}}
\makeatother

\title{\mphubert: Universal Phonetic Inductive Biases for Few-Shot Acoustic Units Discovery}

\author{
  Angelo Ortiz Tandazo$^{\diamond\dagger}$ \quad
  Manel Khentout$^{\diamond}$ \quad
  Youssef Benchekroun$^{\mathsection}$ \\
  \textbf{Thomas Hueber}\thanks{$^{*}$ Equally contributed as senior authors.}$^{\dagger*}$ \quad
  \textbf{Emmanuel Dupoux}$^{\diamond\mathsection*}$
  \\
  \ \\
  $^\diamond$ENS, PSL Research University, EHESS, CNRS, Paris, France
  \\
  $^\dagger$Univ. Grenoble Alpes, CNRS, Grenoble INP, GIPSA-lab, Grenoble, France
  \\
  $^\mathsection$Meta AI Research, France
  \\
  \texttt{angelo.ortiz.tandazo@ens.psl.eu}
}

\begin{document}
\maketitle
\begin{abstract}
This paper introduces \mphubert, a multilingual extension of HuBERT that leverages articulatory features for robust cross-lingual phonetic representation learning.
We continue HuBERT pre-training with supervision based on a phonetic-to-articulatory feature mapping in \num{55} languages.
Our models learn from multilingual data to predict articulatory features or phones, resulting in language-independent representations that capture multilingual phonetic properties.
Through comprehensive ABX discriminability testing, we show \mphubert models produce more context-invariant representations than state-of-the-art multilingual self-supervised learning models.
Additionally, the models effectively adapt to unseen languages and casual speech with minimal self-supervised fine-tuning (\num{10} hours of speech).
This establishes an effective approach for instilling linguistic inductive biases in self-supervised speech models.
\end{abstract}

\section{Introduction}

Is it possible to automatically discover the linguistic units of an unknown language from raw audio only? 
Doing so would be of great help to linguists or speech technologists working on low-resource or unwritten languages \citep{chen2024evaluating, mohamed_22_review_ssl, zelasko22_asrinventories, chen_23_s2s_translation, zhang_21_uwspeech}, or to cognitive modellers trying to understand how children learn their native language before learning to read and write \citep{kuhl_1993_nlm, werker_07_phonetic}.
This question has been addressed using a variety of approaches under the Zero Resource Speech Challenge series \citep{versteegh2015zero, dunbar17_zerospeech, dunbar_22_zrc_lessons}, yielding impressive progress alongside unresolved questions. 

Much of this progress stems from advances in self-supervised learning (SSL) techniques \citep{oord2018representation, baevski2020wav2vec, hsu21_hubert}, which have produced speech representations that capture phonetic structure better than traditional features like MFCCs or mel filterbanks.
This is evidenced by improved discriminability in the learnt representation spaces: two instances of the syllable `bit' lie closer together than one instance of `bit' and one instance of `bet', even across different speakers \citep{schatz16_phdthesis, schatz13_interspeech}.
Further evidence comes from the success of quantisation of these representations, yielding low-bitrate discrete codes suitable for training generative language models that produce novel utterances in the target language \citep{lakhotia2021generative, borsos_23_audiolm, defossez_24_moshi, simon_25_calm}. 

\begin{figure*}[t]
\begin{tikzpicture}[scale=0.7]
    \node at (-3.,2.8) {a.};

    \begin{scope}[xshift=-2.8cm,rotate=90]
        \draw[thick, blue] (-1.1,-0.01) -- (-1.1,0.01);
        \draw[thick, blue] (-1.05,-0.02) -- (-1.05,0.02);
        \draw[thick, blue] (-1.,-0.02) -- (-1.,0.02);
        \draw[thick, blue] (-0.95,-0.03) -- (-0.95,0.03);
        \draw[thick, blue] (-0.9,-0.1) -- (-0.9,0.1);
        \draw[thick, blue] (-0.85,-0.1) -- (-0.85,0.1);
        \draw[thick, blue] (-0.8,-0.15) -- (-0.8,0.15);
        \draw[thick, blue] (-0.75,-0.2) -- (-0.75,0.2);
        \draw[thick, blue] (-0.7,-0.12) -- (-0.7,0.12);
        \draw[thick, blue] (-0.65,-0.1) -- (-0.65,0.1);
        \draw[thick, blue] (-0.6,-0.05) -- (-0.6,0.05);
        \draw[thick, blue] (-0.55,-0.12) -- (-0.55,0.12);
        \draw[thick, blue] (-0.5,-0.15) -- (-0.5,0.15);
        \draw[thick, blue] (-0.45,-0.2) -- (-0.45,0.2);
        \draw[thick, blue] (-0.4,-0.12) -- (-0.4,0.12);
        \draw[thick, blue] (-0.35,-0.2) -- (-0.35,0.2);
        \draw[thick, blue] (-0.3,-0.12) -- (-0.3,0.12);
        \draw[thick, blue] (-0.25,-0.09) -- (-0.25,0.09);
        \draw[thick, blue] (-0.2,-0.07) -- (-0.2,0.07);
        \draw[thick, blue] (-0.15,-0.3) -- (-0.15,0.3);
        \draw[thick, blue] (-0.1,-0.2) -- (-0.1,0.2);
        \draw[thick, blue] (-0.05,-0.2) -- (-0.05,0.2);
        \draw[thick, blue] (0.,-0.15) -- (0.,0.15);
        \draw[thick, blue] (0.05,-0.12) -- (0.05,0.12);
        \draw[thick, blue] (0.1,-0.17) -- (0.1,0.17);
        \draw[thick, blue] (0.15,-0.4) -- (0.15,0.4);
        \draw[thick, blue] (0.2,-0.15) -- (0.2,0.15);
        \draw[thick, blue] (0.25,-0.12) -- (0.25,0.12);
        \draw[thick, blue] (0.3,-0.07) -- (0.3,0.07);
        \draw[thick, blue] (0.35,-0.04) -- (0.35,0.04);
        \draw[thick, blue] (0.4,-0.06) -- (0.4,0.06);
        \draw[thick, blue] (0.45,-0.12) -- (0.45,0.12);
        \draw[thick, blue] (0.5,-0.07) -- (0.5,0.07);
        \draw[thick, blue] (0.55,-0.06) -- (0.55,0.06);
        \draw[thick, blue] (0.6,-0.02) -- (0.6,0.02);
        \draw[thick, blue] (0.65,-0.07) -- (0.65,0.07);
        \draw[thick, blue] (0.7,-0.1) -- (0.7,0.1);
        \draw[thick, blue] (0.75,-0.12) -- (0.75,0.12);
        \draw[thick, blue] (0.8,-0.3) -- (0.8,0.3);
        \draw[thick, blue] (0.85,-0.2) -- (0.85,0.2);
        \draw[thick, blue] (0.9,-0.12) -- (0.9,0.12);
        \draw[thick, blue] (0.95,-0.09) -- (0.95,0.09);
        \draw[thick, blue] (1.0,-0.05) -- (1.,0.05);
        \draw[thick, blue] (1.05,-0.02) -- (1.05,0.02);
        \draw[thick, blue] (1.1,-0.01) -- (1.1,0.01);
    \end{scope}
    
    \node[rectangle,
          minimum width=2.9cm,
          minimum height=2.6cm,
          draw,
          fill=yellow!20,
          rounded corners=0.5mm,
          dashed] (hubert_box) at (-0.07,-0.05) {};
    
    \node[font=\scriptsize,color=gray] at (0.,-1.6) {HuBERT-base};

    \node[trapezium, 
          trapezium left angle=120.5,
          trapezium right angle=120.5,
          trapezium stretches=true,
          minimum width=2.25cm,
          minimum height=0.5cm,
          draw,
          rounded corners=0.5mm,
          fill=lightgray!30,
          align=center,
          font=\tiny\linespread{0.8},
          rotate=90] (feature_extractor) at (-1.4,0) {Feature\\Extractor};
    \node[font=\tiny] at (-1.6,1.1) {\textcolor{cyan}{\SnowflakeChevron}};
    
    \node[rectangle, 
          minimum width=1.6cm,
          minimum height=1.3cm,
          draw,
          rounded corners=0.5mm,
          fill=blue!20] (encoder) at (0.6,0) {};
    \node[font=\tiny] at (0.6,0) {Transformer};
    \node[font=\footnotesize] at (1.5,0.7) {\Fire};
    
    \node at (0.8,1.4) {$\cdots$};
       
    \draw[-{Stealth[length=1.5mm]}, thick] (-2.4,0) -- (feature_extractor.north);
    \draw[-{Stealth[length=1.5mm]}, thick] (feature_extractor.south) -- (encoder.west);

    \draw[dashed,
            fill=orange!20,
            rounded corners=0.5mm] plot coordinates {(-0.9,3) (6.55,3) (6.55,-1.9) (2.2,-1.9) (2.2,1.9) (-0.9,1.9) (-0.9,3)};

    \node[font=\scriptsize,color=gray] at (4.85,2.75) {Downstream Modules};

    \node[rectangle,
          minimum width=2.cm,
          minimum height=0.3cm,
          draw,
          rounded corners=0.5mm,
          font=\tiny,
          fill=red!20] (weighted_sum) at (0.7,2.5) {Weighted Sum};
    \node[font=\footnotesize] at (1.9,2.6) {\Fire};
    
    \draw[-{Stealth[length=1.5mm]}, thick] (-0.3,0.92) -- (-0.3,2.19);
    \draw[-{Stealth[length=1.5mm]}, thick] (0.1,0.92) -- (0.1,2.19);
    \draw[-{Stealth[length=1.5mm]}, thick] (1.5,0.92) -- (1.5,2.19);

    \node[trapezium, 
          trapezium left angle=65.5, 
          trapezium right angle=65.5,
          trapezium stretches=true,
          minimum width=1.8cm,
          minimum height=0.5cm,
          draw,
          rotate=90,
          rounded corners=0.5mm,
          font=\tiny,
          fill=red!20] (projection) at (3.0,0) {Projection};
    \node[font=\footnotesize] at (3.15,0.95) {\Fire};

    \draw[-{Stealth[length=1.5mm]}, thick] (weighted_sum.east) -| (2.35,0) -- (projection.north);

    \node[rectangle,
          minimum width=0.9cm,
          minimum height=1.77cm,
          draw,
          rounded corners=0.5mm,
          font=\tiny,
          fill=red!20] (blstm) at (4.4,0) {BLSTM};
    \node[font=\footnotesize] at (4.9,1.0) {\Fire};
    
    \draw[-{Stealth[length=1.5mm]}, thick] (projection.south) -- (blstm.west);

    \node[trapezium, 
          trapezium left angle=130., 
          trapezium right angle=130.,
          trapezium stretches=true,
          minimum width=2.cm,
          minimum height=0.5cm,
          draw,
          rotate=90,
          align=center,
          rounded corners=0.5mm,
          font=\tiny\linespread{0.8},
          fill=red!20] (feat_projection) at (5.9,0) {Feature\\Proj.};
    \node[font=\footnotesize] at (5.6,1.0) {\Fire};

    \draw[-{Stealth[length=1.5mm]}, thick] (blstm.east) -- (feat_projection.north);

    \node[font=\scriptsize] (pred_feats) at (6.95,0) {$\mathbf{\tilde{f}}$};

    \draw[-, thick] (feat_projection.south) -- (pred_feats.west);
    
    \node[rectangle,
          minimum width=0.4cm,
          minimum height=0.6cm,
          draw,
          font=\scriptsize,
          rounded corners=0.5mm] (stop_gradient) at (7.95,0) {sg};
    
    \draw[-{Stealth[length=1.5mm]}, thick] (pred_feats.east) -- (stop_gradient.west);

    \node[trapezium, 
          trapezium left angle=38, 
          trapezium right angle=38,
          trapezium stretches=true,
          minimum width=2.6cm,
          minimum height=0.5cm,
          draw,
          rotate=90,
          rounded corners=0.5mm,
          align=center,
          font=\tiny\linespread{0.8},
          fill=olive!20] (mlp) at (9.1,0) {Phone\\Model};
    \node[font=\footnotesize] at (9.35,1.1) {\Fire};

    \draw[-{Stealth[length=1.5mm]}, thick] (stop_gradient.east) -- (mlp);

    \node[font=\scriptsize] (pred_phones) at (10.1,0) {$\mathbf{\tilde{p}}$};

    \draw[-, thick] (mlp.south) -- (pred_phones.west);
    
    \node[rectangle,
          minimum width=2.cm,
          minimum height=1cm,
          draw,
          fill=lightgray!20,
          rounded corners=0.5mm,
          align=center,
          dashed
    ] (gt_b) at (3.5,-3.3) {};
    \node[align=center,color=gray,font=\tiny] at (3.1,-3.) {Gold features};
    \node[font=\scriptsize] (gt_feats) at (4.3,-3) {$\mathbf{f}$};
    \node[align=center,color=gray,font=\tiny] at (3.06,-3.6) {Gold phones};
    \node[font=\scriptsize] (gt_phones) at (4.3,-3.6) {$\mathbf{p}$};
    \node[font=\tiny] (bce_b) at (6.95,-3) {Cross-entropy};
    \node[font=\tiny] (ce_b) at (10.1,-3.6) {Cross-entropy};

    \draw[-{Stealth[length=1.5mm]}, thick] (pred_feats.south) -- (bce_b.north);
    \draw[-{Stealth[length=1.5mm]}, thick] (gt_feats.east) -- (bce_b.west);
    \draw[-{Stealth[length=1.5mm]}, thick] (pred_phones.south) -- (ce_b.north);
    \draw[-{Stealth[length=1.5mm]}, thick] (gt_phones.east) -- (ce_b.west);

    
    \node at (11.3,2.8) {b.};

    \begin{scope}[xshift=11.6cm,yshift=-0.6cm,rotate=90]
        \draw[thick, blue] (-1.1,-0.01) -- (-1.1,0.01);
        \draw[thick, blue] (-1.05,-0.02) -- (-1.05,0.02);
        \draw[thick, blue] (-1.,-0.02) -- (-1.,0.02);
        \draw[thick, blue] (-0.95,-0.03) -- (-0.95,0.03);
        \draw[thick, blue] (-0.9,-0.1) -- (-0.9,0.1);
        \draw[thick, blue] (-0.85,-0.1) -- (-0.85,0.1);
        \draw[thick, blue] (-0.8,-0.15) -- (-0.8,0.15);
        \draw[thick, blue] (-0.75,-0.2) -- (-0.75,0.2);
        \draw[thick, blue] (-0.7,-0.12) -- (-0.7,0.12);
        \draw[thick, blue] (-0.65,-0.1) -- (-0.65,0.1);
        \draw[thick, blue] (-0.6,-0.05) -- (-0.6,0.05);
        \draw[thick, blue] (-0.55,-0.12) -- (-0.55,0.12);
        \draw[thick, blue] (-0.5,-0.15) -- (-0.5,0.15);
        \draw[thick, blue] (-0.45,-0.2) -- (-0.45,0.2);
        \draw[thick, blue] (-0.4,-0.12) -- (-0.4,0.12);
        \draw[thick, blue] (-0.35,-0.2) -- (-0.35,0.2);
        \draw[thick, blue] (-0.3,-0.12) -- (-0.3,0.12);
        \draw[thick, blue] (-0.25,-0.09) -- (-0.25,0.09);
        \draw[thick, blue] (-0.2,-0.07) -- (-0.2,0.07);
        \draw[thick, blue] (-0.15,-0.3) -- (-0.15,0.3);
        \draw[thick, blue] (-0.1,-0.2) -- (-0.1,0.2);
        \draw[thick, blue] (-0.05,-0.2) -- (-0.05,0.2);
        \draw[thick, blue] (0.,-0.15) -- (0.,0.15);
        \draw[thick, blue] (0.05,-0.12) -- (0.05,0.12);
        \draw[thick, blue] (0.1,-0.17) -- (0.1,0.17);
        \draw[thick, blue] (0.15,-0.4) -- (0.15,0.4);
        \draw[thick, blue] (0.2,-0.15) -- (0.2,0.15);
        \draw[thick, blue] (0.25,-0.12) -- (0.25,0.12);
        \draw[thick, blue] (0.3,-0.07) -- (0.3,0.07);
        \draw[thick, blue] (0.35,-0.04) -- (0.35,0.04);
        \draw[thick, blue] (0.4,-0.06) -- (0.4,0.06);
        \draw[thick, blue] (0.45,-0.12) -- (0.45,0.12);
        \draw[thick, blue] (0.5,-0.07) -- (0.5,0.07);
        \draw[thick, blue] (0.55,-0.06) -- (0.55,0.06);
        \draw[thick, blue] (0.6,-0.02) -- (0.6,0.02);
        \draw[thick, blue] (0.65,-0.07) -- (0.65,0.07);
        \draw[thick, blue] (0.7,-0.1) -- (0.7,0.1);
        \draw[thick, blue] (0.75,-0.12) -- (0.75,0.12);
        \draw[thick, blue] (0.8,-0.3) -- (0.8,0.3);
        \draw[thick, blue] (0.85,-0.2) -- (0.85,0.2);
        \draw[thick, blue] (0.9,-0.12) -- (0.9,0.12);
        \draw[thick, blue] (0.95,-0.09) -- (0.95,0.09);
        \draw[thick, blue] (1.0,-0.05) -- (1.,0.05);
        \draw[thick, blue] (1.05,-0.02) -- (1.05,0.02);
        \draw[thick, blue] (1.1,-0.01) -- (1.1,0.01);
    \end{scope}
    
    \node[font=\scriptsize,color=black] (pred_d) at (15.7,3.) {Teacher};

    \node[rectangle,
          minimum width=0.6cm,
          minimum height=0.6cm,
          draw,
          fill=yellow!20,
          rounded corners=0.5mm,
          font=\tiny
    ] (hubert_c) at (13.1,1.5) {Hb};
    \node[font=\tiny] at (13.35,1.75) {\textcolor{cyan}{\SnowflakeChevron}};
    
    \node[rectangle,
          minimum width=0.6cm,
          minimum height=0.6cm,
          draw,
          fill=orange!20,
          rounded corners=0.5mm,
          align=center,
          font=\tiny
    ] (down_c) at (14.5,1.5) {DM};
    \node[font=\tiny] at (14.75,1.75) {\textcolor{cyan}{\SnowflakeChevron}};

    \draw[-, thick] (12.3,1.52) -- (12.3,-2.72);
    \draw[-, thick] (12.0,-0.6) -- (12.3,-0.6);
    \draw[-{Stealth[length=1.5mm]}, thick] (12.3,1.5) -- (hubert_c.west);
    \draw[-{Stealth[length=1.5mm]}, thick] (hubert_c.east) -- (down_c.west);
    \draw[-{Stealth[length=1.5mm]}, thick] (down_c.east) -- (15.4,1.5);

    \begin{axis}[
      xshift=16.cm,
      yshift=1.cm,
      no markers, 
      domain=0:7,
      samples=150,
      ymin=0,
      ymax=1,
      axis lines*=left, 
      height=3cm, 
      width=4cm,
      font=\tiny,
      xtick=\empty, 
      ytick=\empty,
      enlargelimits=false, 
      axis on top,
      clip=false
    ]
    
    \addplot[red!50!black,domain=0.2:7,name path=curve,restrict y to domain=-inf:1]   
      {1/(4*x)};
    \addplot[name path=xaxis] 
      {0};
    
    \addplot[orange!30] 
      fill between[of=curve and xaxis,soft clip={domain=0.2:3}];
    
    \draw[gray,dashed] 
      (axis cs:3,0) -- (axis cs:3,1);
    
    \node[right,align=left,anchor=north west] at (axis cs:3,1.15)  {Threshold}; 
    
    \coordinate (aux2) at (axis cs:2,{1/(4*2)});
    \node[align=center,anchor=south west] 
      at ([xshift=0.5cm,yshift=-5pt]aux2)
      (sig)
      {Target clusters \\ $\{ \mathbf{c_k} \}$};
    \draw
      (sig.west) -- ([yshift=-2pt]aux2);
    
    \node[anchor=south west,align=left,rotate=90] 
      at (axis cs:0.3,-0.1)
      {Frequency};
    \node[anchor=north west,align=center] 
      at (axis cs:0,0)
      {Feature vectors by \\ frequency rank};
    \end{axis}    
    
    \node[font=\scriptsize,color=black] (pred_d) at (15.7,-4.8) {Student};

    \node[rectangle,
          minimum width=3.75cm,
          minimum height=1.8cm,
          draw,
          fill=yellow!20,
          rounded corners=0.5mm,
          dashed] (hubert_box_d) at (15.13,-2.7) {};
    
    \node[font=\tiny,color=gray] (pred_d) at (15.11,-3.8) {HuBERT-base};

    \node[trapezium, 
          trapezium left angle=120.5,
          trapezium right angle=120.5,
          trapezium stretches=true,
          minimum width=1.25cm,
          minimum height=0.5cm,
          draw,
          rounded corners=0.5mm,
          fill=lightgray!30,
          align=center,
          font=\tiny,
          rotate=90] (feature_extractor_d) at (13.1,-2.7) {FE};
    \node[font=\footnotesize] at (13.0,-2.3) {\Fire};

    \node[rectangle, 
          minimum width=0.3cm,
          minimum height=1.cm,
          draw,
          fill=teal!40] (unmasked_d) at (14.1,-2.7) {};

    \node[font=\scriptsize] () at (14.1,-1.7) {$\mathbf{x}$};
    
    \node[rectangle, 
          minimum width=0.3cm,
          minimum height=1.cm,
          draw,
          fill=teal!40] (masked_d) at (15.1,-2.7) {};

    \node[font=\scriptsize] () at (15.1,-1.66) {$\mathbf{\tilde{x}}$};

    \draw[black] (14.88,-2.05) -- (15.32,-2.05);
    \draw[black] (14.88,-2.07) -- (15.32,-2.07);
    \draw[black] (14.88,-2.17) -- (15.32,-2.17);
    \draw[black] (14.88,-2.19) -- (15.32,-2.19);
    \draw[black] (14.88,-2.21) -- (15.32,-2.21);
    \draw[black] (14.88,-2.25) -- (15.32,-2.25);
    \draw[black] (14.88,-2.27) -- (15.32,-2.27);
    \draw[black] (14.88,-2.37) -- (15.32,-2.37);
    \draw[black] (14.88,-2.41) -- (15.32,-2.41);
    \draw[black] (14.88,-2.43) -- (15.32,-2.43);
    \draw[black] (14.88,-2.45) -- (15.32,-2.45);
    \draw[black] (14.88,-2.47) -- (15.32,-2.47);
    \draw[black] (14.88,-2.49) -- (15.32,-2.49);
    \draw[black] (14.88,-2.57) -- (15.32,-2.57);
    \draw[black] (14.88,-2.60) -- (15.32,-2.60);
    \draw[black] (14.88,-2.64) -- (15.32,-2.64);
    \draw[black] (14.88,-2.66) -- (15.32,-2.66);
    \draw[black] (14.88,-2.72) -- (15.32,-2.72);
    \draw[black] (14.88,-2.77) -- (15.32,-2.77);
    \draw[black] (14.88,-2.79) -- (15.32,-2.79);
    \draw[black] (14.88,-2.81) -- (15.32,-2.81);
    \draw[black] (14.88,-2.83) -- (15.32,-2.83);
    \draw[black] (14.88,-2.85) -- (15.32,-2.85);
    \draw[black] (14.88,-2.87) -- (15.32,-2.87);
    \draw[black] (14.88,-2.92) -- (15.32,-2.92);
    \draw[black] (14.88,-2.94) -- (15.32,-2.94);
    \draw[black] (14.88,-2.97) -- (15.32,-2.97);
    \draw[black] (14.88,-3.01) -- (15.32,-3.01);
    \draw[black] (14.88,-3.07) -- (15.32,-3.07);
    \draw[black] (14.88,-3.14) -- (15.32,-3.14);
    \draw[black] (14.88,-3.16) -- (15.32,-3.16);
    \draw[black] (14.88,-3.18) -- (15.32,-3.18);
    \draw[black] (14.88,-3.23) -- (15.32,-3.23);
    \draw[black] (14.88,-3.25) -- (15.32,-3.25);
    \draw[black] (14.88,-3.28) -- (15.32,-3.28);
    \draw[black] (14.88,-3.33) -- (15.32,-3.33);
    \draw[black] (14.88,-3.35) -- (15.32,-3.35);

    \node[rectangle, 
          minimum width=1.cm,
          minimum height=0.7cm,
          draw,
          rounded corners=0.5mm,
          font=\tiny,
          fill=blue!20] (transformer_d) at (16.7,-2.7) {Transformer};
    \node[font=\footnotesize] at (17.4,-2.4) {\Fire};

    \node[font=\scriptsize] (pred_d) at (18.3,-2.7) {$\mathbf{\tilde{z}}$};
    
    \node[font=\tiny] (ce_d) at (18.5,-4.5) {Cross-entropy};
    
    TODO: resume here!
    \draw[-{Stealth[length=1.5mm]}, thick] (12.3,-2.7) -- (feature_extractor_d.north);
    \draw[-{Stealth[length=1.5mm]}, thick] (feature_extractor_d.south) -- (unmasked_d.west);
    \draw[-{Stealth[length=1.5mm]}, thick] (unmasked_d.east) -- (masked_d.west);
    \draw[-{Stealth[length=1.5mm]}, thick] (masked_d.east) -- (transformer_d.west);
    \draw[-, thick] (transformer_d.east) -- (pred_d.west);
    \draw[-{Stealth[length=1.5mm]}, thick] (pred_d.south) -- (18.3,-4.2);
    \draw[-{Stealth[length=1.5mm]}, thick] (18.8,1.4) -- (18.8,-4.2);
\end{tikzpicture}
\caption{\textbf{a. Multilingual training.} \mphubert-feat is trained to recognise ternary-valued articulatory features and phones using an encoder (HuBERT-base), downstream modules (weighted sum, up-projection, two-layer BLSTM, feature projection), and a phone model (two-layer perceptron); the feature states receive no gradients from the phone recognition loss due to the stop-gradient operator (\texttt{sg}).
\textbf{b. Self-supervised fine-tuning.} \textit{Top:} Offline clustering is applied to one of the layers of \mphubert, the teacher network, on an unseen language; \textit{bottom:} the \mphubert encoder, the student network, is then trained to predict the corresponding clusters of masked input.}
\label{fig:model}
\vspace{-0.5em}
\end{figure*}

However, current approaches face two limitations.
First, units discovered through speech SSL do not correspond one-to-one with linguistic units like phones, syllables or words.
After clustering, these units are typically shorter and more numerous than standard linguistic units: \qtyrange[range-phrase = --, range-units = single]{20}{40}{\milli\second} long vs.~\qty{70}{\milli\second} for phonemes, and $ N =$ \numrange[range-phrase = --]{100}{1000} vs.~\numrange[range-phrase = --]{10}{100} for phonemes \citep{lavechin_25_simulating, schatz_21_pnas}.
Moreover, they lack full invariance to speaker identity \citep{deseyssel22_probing, mohamed24_interspeech} and phonetic context \citep{hallap23_any-ctx}, suggesting they capture acoustic events rather than abstract linguistic units.
As a result, they produce codes with higher bitrates than phonemic transcriptions: about \qtyrange[range-phrase = --, range-units = single, per-mode = symbol]{100}{150}{\bit\per\second} versus \qtyrange[range-phrase = --, range-units = single, per-mode = symbol]{50}{70}{\bit\per\second} \citep{lakhotia2021generative, dunbar_22_zrc_lessons}.
Second, current SSL algorithms require massive amounts of clean speech: \citet{hsu21_hubert} uses \num{960} hours of clean English audio, \citet{zanonboito24_mhubert} uses \qty{90}{\kilo\relax} hours, and \citet{chen24_xeus} uses \qty{1}{\mega\relax} hours.
Such quantities are unavailable for low-resource languages, and notably, children acquire their language's phonetics with far less than \qty{1000}{\hour} of much noisier input.

One avenue for improving SSL models involves pre-training universal models \citep{conneau2020unsupervised}, with recent work expanding both language coverage and training data \citep{babu22_xls-r, zanonboito24_mhubert, pratap24_mms, chen24_xeus}.
Inspired by the International Phonetic Alphabet (IPA), another research direction explores how phonetically-informed target signals influence learnt representations and their cross-lingual transferability \citep{wang22m_interspeech, ma23b_interspeech, feng23_interspeech}, suggesting that explicit phonological supervision enhances speech models' cross-lingual capabilities.

In this paper, we explore the hypothesis that standard SSL algorithms lack \textbf{strong inductive biases} necessary for learning invariant speech representations from limited audio data in new languages.
Following the universal pre-training and phonetically-informed research lines, we propose transforming a monolingual pre-trained SSL model (specifically, HuBERT-base trained on English) into a universal SSL model with strong inductive biases by fine-tuning it on \textbf{universal IPA phonemes and features} across \num{55} diverse languages.
By directly addressing the limitations of prior work, such as the need for large datasets and the absence of explicit phonological supervision, our method provides a practical and linguistically informed solution for both speech technology and language documentation in low-resource settings.
We evaluate this model, coined \mphubert, on the ZRC\num{2017} challenge, which presents 5 languages with less than \qty{10}{\hour} of training data (English, French, German, Mandarin, Wolof).
To increase the evaluation's diversity and validity, we extend the ZRC\num{2017} benchmark with \num{5} typologically diverse languages (Swahili, Tamil, Thai, Turkish, Ukrainian).
Evaluation employs the within- and across-speaker ABX metrics from ZRC\num{2017}, supplemented with metrics measuring invariance to contextual allophony \citep{hallap23_any-ctx}.

Our main contributions are twofold:
\begin{enumerate*}[label=(\roman*)]
    \item We demonstrate that multilingual supervised fine-tuning of HuBERT for articulatory feature or phone prediction creates robust multilingual phonetic representations with strong zero-shot transfer capabilities.
    \item Our resulting models enable effective adaptation to unseen languages and casual speech with minimal self-supervised fine-tuning, achieving strong speaker and contextual invariance in new languages with only \qty{10}{\hour} of unlabelled data.
\end{enumerate*}
As a by-product, our method also yields candidate phoneme and feature sets for unseen languages, with potential applications for linguistic analyses of low-resource languages.
The code with the data processing, methods and baselines can be found at
\ifaclpubformat
\url{https://github.com/bootphon/maubert}.
\else
[to ensure the anonymity of the author, the link to the resource will be added after the review process].
\fi

\section{Related Work}
\vspace{-0.1em}

\paragraph{Multilingual Speech Representation Learning.}
The field of multilingual speech processing has grown rapidly with large-scale semi- or self-supervised learning models that showed the potential for cross-lingual representation learning with little to no supervision \citep{wang2021unispeech, conneau2020unsupervised}.
Recent studies have expanded language coverage \citep{babu22_xls-r}, diversified data sources \citep{pratap24_mms}, and improved efficiency \citep{zanonboito24_mhubert} and robustness to noise \citep{chen24_xeus}.
These multilingual SSL models build upon foundational work in self-supervised speech representation learning \citep{baevski2020wav2vec, hsu21_hubert, chen_22_wavlm} and have been evaluated with multi-task frameworks like SUPERB \citep{yang21c_superb, shi23g_ml_superb}.
Meanwhile, other studies have explored the impact of phonetically-informed targets on learnt representations and their cross-lingual transferability \citep{wang22m_interspeech, ma23b_interspeech, feng23_interspeech}.
Inspired by the downstream framework of SUPERB, this work extends HuBERT-base for articulatory feature prediction.

\paragraph{Articulatory Features in Speech Processing.}
Early work established the use of articulatory features (AFs) in speech processing \citep{deng_91_hmm, elenius91_eurospeech, eide_93_lingfeats}, demonstrated that supervised learning can be used to automatically extract phonological features from raw (and continuous) speech \citep{papcun1992inferring, king00_phonfeats_rnn} and produced robust articulatory/phonological feature-based speech technologies \citep{kirchhoff1999robust, livescu2007articulatory, frankel2007articulatory}.
The development of systematic feature inventories, particularly PanPhon \citep{mortensen16_panphon}, has provided practical computational tools for cross-linguistic analysis.
This has enabled recent efforts to explore AFs in multilingual contexts, demonstrating their effectiveness for zero-shot multilingual speech synthesis \citep{staib20_interspeech} or showing their utility for cross-lingual speech recognition in low-resource languages \citep{feng23_interspeech}.

\paragraph{Evaluation of Speech Representations.}
More specific subtasks have been developed as alternatives to downstream-based evaluation, offering clearer insights into unsupervised language learning.
A prominent example is the ABX discriminability evaluation \citep{schatz16_phdthesis}, which assesses whether learnt representations can distinguish between different phonetic units in a way that reflects human perceptual boundaries.
The Zero Resource Speech Challenge series \citep{versteegh2015zero, dunbar17_zerospeech} has systematically applied ABX evaluation to assess unsupervised speech representations, establishing benchmarks for phonetic discrimination across diverse languages and speakers.
While ABX testing shows sufficient correlation with downstream performance to serve as a model comparison proxy, traditional ABX evaluation has not assessed other types of invariance, like speaking rate or speech style variations \citep{dunbar_22_zrc_lessons}.
A recent extension has begun addressing this limitation by measuring context invariance \citep{hallap23_any-ctx}.
The present work builds on this extension and adds the comparison between read and casual speech.

\section{\mphubert}

In this section, we introduce our \textbf{M}ultilingual \textbf{a}rticulatory hidden-\textbf{u}nit \textbf{BERT} (\mphubert) models (\Cref{fig:model}).
We describe the base architecture for multilingual training (\S \ref{sec:archi}), and the self-supervised fine-tuning approach (\S \ref{sec:ssl}).

\subsection{Multilingual Pre-Training}
\label{sec:archi}

\mphubert models are based on multilingual, continual learning of a pre-trained self-supervised speech model for articulatory feature (AF) or phone recognition (\Cref{fig:model}a).
We re-train HuBERT \citep{hsu21_hubert} using the VoxCommunis Corpus \citep{ahn22_voxcommunis}, and the associated featural annotations extracted with PanPhon \citep{mortensen16_panphon}.

We propose two versions of \mphubert: \feat and \phone.
The former incorporates an AF bottleneck (\Cref{fig:model}a), while the latter directly predicts phones without intermediate AFs\footnote{The feature projection in \Cref{fig:model}a is replaced with a phone projection, and the phone model is dropped.}. 

\paragraph{Encoder.}
We use the pre-trained HuBERT-base model as our \textit{encoder}.
The convolutional feature extractor is kept frozen, but the Transformer encoder is trainable.
We extract the feature extractor's output after layer normalisation and dropout, as well as the outputs from each of the \num{12} Transformer encoder layers.
The input masking is disabled during continual pre-training following SUPERB's downstream framework \citep{yang21c_superb}.

\paragraph{Downstream modules.}
Preliminary experiments revealed that a simple linear layer on top of HuBERT was insufficient for the feature recognition task. Following the SUPERB framework for ASR \citep{yang21c_superb}, we instead leverage a weighted sum of HuBERT's intermediate representations fed into a BLSTM. Ablation experiments confirmed the importance of this contextual architecture: replacing the BLSTM with non-contextual networks consistently degraded performance on both the multilingual recognition tasks and the phonetic probing, motivating our adoption of the SUPERB design.

Concretely, we first compute a weighted sum of the intermediate representations from our encoder and up-project them to a \num{1024}-dimensional space.
These representations are then processed through a bidirectional two-layer LSTM.
Finally, we down-project the concatenated forward and backward output states into task-specific spaces: a \num{22}-dimensional AF space for \mphubert-\feat and a \num{3293}-dimensional phone space for \mphubert-\phone.

\begin{table}[t]
  \centering
  \resizebox{0.47\textwidth}{!}{%
      \begin{tabular}{ l l c c c }
        \toprule
        \multirow{2}{1cm}{Eval. lang.} & \multirow{2}{2cm}{\mphubert variant} & \multirow{2}{1cm}{Feat. acc. $\uparrow$} & \multirow{2}{1cm}{Phone acc. $\uparrow$} & \multirow{2}{1cm}{PER $\downarrow$} \\
        \\
        \midrule
        \multirow{2}{*}{Train} & \feat & \textbf{\num[text-series-to-math]{95.60}} & \num{82.35} & \num{30.64} \\ %
        & \phone & \num{92.72} & \textbf{\num[text-series-to-math]{82.72}} & \textbf{\num[text-series-to-math]{28.69}} \\
        \midrule
        \multirow{2}{*}{Dev} & \feat & \textbf{\num[text-series-to-math]{92.35}} & \num{66.74} & \num{50.46} \\
        & \phone & \num{88.57} & \textbf{\num[text-series-to-math]{67.15}} & \textbf{\num[text-series-to-math]{48.38}} \\
        \bottomrule
      \end{tabular}%
  }
  \caption{Feature and phone evaluation of \mphubert on the held-out test set of the \num{55} training languages and zero-shot performance on the \num{5} development languages. All scores are in \%.}
  \label{tab:feat-phone-acc}

  \vspace{1em}

  \includegraphics[width=0.47\textwidth]{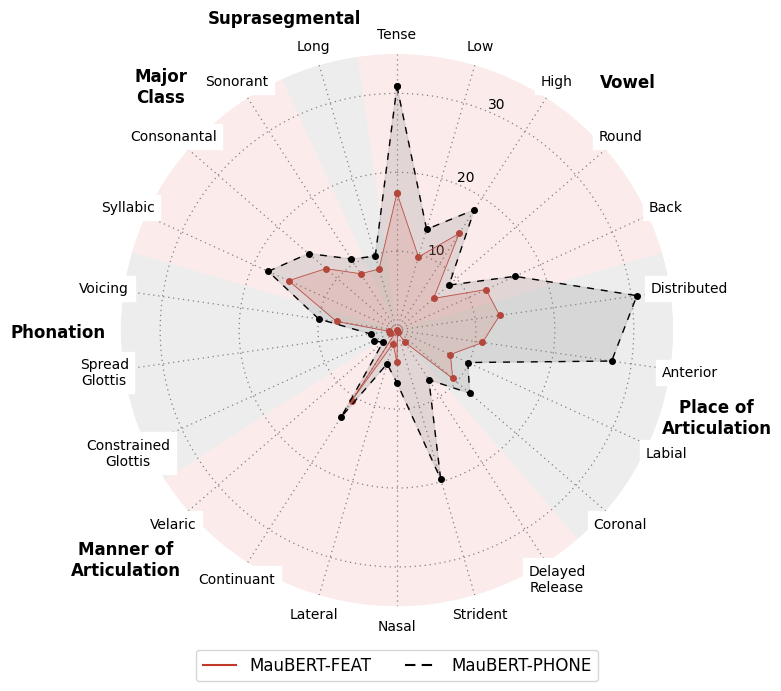}
  \captionof{figure}{Classification errors across articulatory features ($\downarrow$) for both \mphubert variants on the \num{5} development languages. All values are in \%.}
  \label{fig:radar_fr}
\end{table}

\paragraph{Phone model.}
Given the non-injective nature of the feature-to-phone mapping, for \mphubert-\feat, we jointly learn a phone model consisting of a two-layer perceptron.
Since we want the pre-training to be led by the feature recognition task only, a stop gradient operator prevents the feature hidden states from receiving any gradients from the phone recognition loss.

\subsection{Self-Supervised Fine-Tuning}
\label{sec:ssl}

We employ self-supervised fine-tuning to adapt \mphubert models to unseen languages with limited or no labelled data.
This approach generates pseudo-labels through clustering of learnt representations and applies masked language modelling (\Cref{fig:model}b), enabling \mphubert to adapt to the acoustic patterns of new languages.

We use four methods to generate pseudo-labels: K-means, frequent features, frequent phones and all phones.
As \citet{hsu21_hubert}, we apply K-means clustering with $K = 100$ to representations from encoder Transformer layers (HuBERT-base, \mphubert) or downstream module layers (\mphubert variants).
For \mphubert-\feat, we extract the top $K$ most frequent feature vectors (\textit{feat. freq.}) from the articulatory feature space (top of \Cref{fig:model}b).
For both \mphubert variants, we extract the top $K$ most frequent phones (\textit{phone freq.}) or all phones from pre-training data (\textit{all phones}).
See \Cref{app:clustering} for details.

\begin{table}[t]
    \centering
    \resizebox{0.47\textwidth}{!}{%
        \begin{tabular}{ l r r r c c}
            \toprule
            \multirow{2}{2cm}{Model} & \multirow{2}{1.6cm}{\# Params} & \multirow{2}{1.2cm}{\# Langs} & \multirow{2}{1.2cm}{\# Hours} & Seen & Seen \\
            & & & & dev & test \\
            \midrule
            MMS & \qty{965}{\mega\relax} & \num{1406} & \qty{491}{\kilo\relax} & \num{5} & \num{5} \\
            XEUS & \qty{577}{\mega\relax}  & \num{4057} & \qty{1}{\mega\relax} & \num{5} & \num{5} \\
            mHuBERT-\num{147} & \qty{95}{\mega\relax}  & \num{147} & \qty{90}{\kilo\relax} & \num{5} & \num{4} \\
            HuBERT-base & \qty{95}{\mega\relax}  & \num{1} & \num{960} & \num{0} & \num{1} \\
            \midrule
            \mphubert (ours) & \qtyrange[ range-phrase = --, range-units = single]{141}{144}{\mega\relax}  & \num{55} & \num{788} & \num{0} & \num{1}\tablefootnote{The backbone of our models being HuBERT-base, some English influence might remain in our models' weights.} \\
            \bottomrule
        \end{tabular}%
    }
    \caption{Comparison of speech models by number of parameters, number of languages, training data size, and development and test languages seen during training (continual learning for \mphubert).}
    \label{tab:baselines}
\end{table}

\section{\mphubert Multilingual Pre-Training}

This section describes the multilingual training and evaluation of \mphubert variants for articulatory feature and phone recognition.

\subsection{Data Processing}

We use the VoxCommunis Corpus, which provides phone-level annotations for a subset of Common Voice \citep{ardila20_common_voice} obtained with Montreal Forced Aligner \citep{mcauliffe17_mfa}.
Of the \num{63} covered languages, \num{55} are used for supervised articulatory feature prediction (totalling \qty{788.4}{\hour} hours), \num{5} serve as development languages, and \num{3} are discarded as they were test languages.
(Refer to \S \ref{sec:abx_data} for the development and test languages and to \Cref{app:data} for more data processing details.)

Using PanPhon's feature table\footnote{We exclude PanPhon's two tonal features from the \num{24} AFs since VoxCommunis alignments lack tone segments.}, ternary-feature\footnote{Features take `+', `-' or `\num{0}' values, with zero indicating context-dependent values (\textit{e.g.} \texttt{high} for \textipa{[R]}) or irrelevance to the phone (\textit{e.g.} \texttt{strident} for vowels).} annotations are derived from the phone-level annotations.
Annotated segments incompatible with PanPhon are manually fixed (\textit{e.g.} \textipa{[tS]} $\rightarrow$ \textipa{[\t{tS}]}, \textipa{[b\super H]} $\rightarrow$ \textipa{[\"*b]}, \textipa{[\textg]} $\rightarrow$ \textipa{[g]}).
Finally, we collapse the IPA table by keeping only distinct feature \textit{vectors} (\textit{e.g.} \textipa{[\"\ae]}, \textipa{[e\super Q]}, \textipa{[5\super Q]}, \textipa{[\|`\ae]} and \textipa{[\ae]} are all represented by the same feature vector), which reduces the table size from \num{6367} to \num{3293} segment representatives.
These representatives are then used for both phone recognition 
and feature recognition (underlying feature values).

\begin{table*}[t]
  \centering
  \resizebox{\textwidth}{!}{%
    \addtolength{\tabcolsep}{-0.2em}
    \begin{tabular}{ l c c c c c c c c c c c c >{\columncolor[gray]{0.9}}c c c c c c c c c c c >{\columncolor[gray]{0.9}}c }
    \toprule
    \multicolumn{3}{c}{Systems} & & \multicolumn{10}{c}{Development languages} & & \multicolumn{10}{c}{Test languages (ZRC\num{2017})} \\
    \cmidrule{1-25}
    \multirow{3}{*}{Model} & \multirow{3}{*}{Layer} & \multirow{3}{*}{\# units} & & \multicolumn{2}{c}{triphone} & & \multicolumn{5}{c}{phoneme ABX $\downarrow$} & & & & \multicolumn{8}{c}{triphone ABX $\downarrow$} & & \\
    & & & & \multicolumn{2}{c}{ABX $\downarrow$} & & \multicolumn{2}{c}{within ctx} & & \multicolumn{2}{c}{any ctx} & & \multirow{-2}{*}{\textit{avg.}} & & \multicolumn{2}{c}{\qty{1}{\second}} & & \multicolumn{2}{c}{\qty{10}{\second}} & & \multicolumn{2}{c}{\qty{120}{\second}} & & \multirow{-2}{*}{\textit{avg.}} \\
    \cmidrule{5-6} \cmidrule{8-9} \cmidrule{11-12} \cmidrule{16-17} \cmidrule{19-20} \cmidrule{22-23}
    & & & & WS & AS & & WS & AS & & WS & AS & & \cellcolor{white} & & WS & AS & & WS & AS & & WS & AS \\
    \midrule
    \quad\textit{Zero-shot} \\
    MFCC & - & \num{39} & & \num{20.00} & \num{29.00} & & \num{13.23} & \num{22.36} & & \num{18.05} & \num{26.33} & & \num{21.49} & & \num{14.78} & \num{25.58} & & \num{14.70} & \num{25.33} & & \num{14.70} & \num{25.32} & & \num{20.07} \\
    MMS-\num{1}B & \num{34} & \num{1280} & & \num{9.37} & \num{10.74} & & \num{4.76} & \num{6.02} & & \num{10.53} & \num{11.37} & & \num{8.80} & & \num{7.58} & \num{9.02} & & \num{6.91} & \num{7.91} & & \num{6.91} & \num{7.83} & & \num{7.69} \\
    XEUS & \num{18} & \num{1024} & & \num{6.14} & \num{7.15} & & \num{3.58} & \num{4.52} & & \num{9.28} & \num{9.45} & & \num{6.69} & & \textbf{\num[text-series-to-math]{4.67}} & \textbf{\num[text-series-to-math]{5.68}} & & \textbf{\num[text-series-to-math]{4.19}} & \textbf{\num[text-series-to-math]{4.91}} & & \textbf{\num[text-series-to-math]{4.29}} & \textbf{\num[text-series-to-math]{4.99}} & & \textbf{\num[text-series-to-math]{4.79}} \\
    mHuBERT-\num{147} & 
\num{7} & \num{768} & & \num{7.37} & \num{8.64} & & \num{3.70} & \num{4.80} & & \num{9.00} & \num{9.51} & & \num{7.17} & & \num{6.93} & \num{8.13} & & \num{5.75} & \num{6.49} & & \num{6.67} & \num{7.78} & & \num{6.96} \\
    HuBERT-base & \num{11} & \num{768} & & \num{6.77} & \num{8.18} & & \num{3.77} & \num{4.92} & & \num{8.55} & \num{9.19} & & \num{6.90} & & \num{6.21} & \num{7.42} & & \num{5.31} & \num{6.21} & & \num{5.62} & \num{6.62} & & \num{6.23} \\
    \cdashline{1-25}
    \mphubert \\
    \quad \feat & \num{9} & \num{768} & & \underline{\num{5.49}} & \underline{\num{6.52}} & & \textbf{\num[text-series-to-math]{2.95}} & \underline{\num{3.81}} & & \underline{\num{5.97}} & \underline{\num{6.47}} & & \underline{\num{5.20}} & & \num{5.86} & \num{6.84} & & \num{4.78} & \underline{\num{5.57}} & & \num{4.86} & \num{5.68} & & \num{5.60} \\
    \quad \phone & proj & \num{1024} & & \textbf{\num[text-series-to-math]{5.42}} & \textbf{\num[text-series-to-math]{6.46}} & & \underline{\num{2.96}} & \textbf{\num[text-series-to-math]{3.79}} & & \textbf{\num[text-series-to-math]{5.49}} & \textbf{\num[text-series-to-math]{6.12}} & & \textbf{\num[text-series-to-math]{5.04}} & & \underline{\num{5.36}} & \underline{\num{6.44}} & & \underline{\num{4.68}} & \num{5.58} & & \underline{\num{4.68}} & \underline{\num{5.60}} & & \underline{\num{5.39}} \\
    \midrule
    \quad\textit{supervised FT (10 h)} \\
    HuBERT-base \\
    \quad + PR & ws & \num{768} & & \num{4.87} & \num{6.13} & & \num{2.30} & \num{3.09} & & \num{3.65} & \num{4.17} & & \num{4.04} & & \num{5.52} & \num{6.67} & & \num{4.10} & \num{4.99} & & \num{4.51} & \num{5.49} & & \num{5.21} \\
    \quad + MPR & \num{11} & \num{768} & & \num{4.26} & \num{4.98} & & \num{2.05} & \num{2.62} & & \num{3.94} & \num{4.30} & & \num{3.69} & & \num{4.26} & \num{4.84} & & \num{3.25} & \num{3.73} & & \num{3.89} & \num{4.36} & & \num{4.05} \\
    \cdashline{1-25}
    \mphubert \\
    \quad \feat + MPR & \num{11} & \num{768} & & \underline{\num{3.65}} & \textbf{\num[text-series-to-math]{4.38}} & & \underline{\num{1.83}} & \textbf{\num[text-series-to-math]{2.28}} & & \underline{\num{3.17}} & \underline{\num{3.44}} & & \underline{\num{3.13}} & & \textbf{\num[text-series-to-math]{3.81}} & \textbf{\num[text-series-to-math]{4.26}} & & \underline{\num{2.86}} & \underline{\num{3.25}} & & \underline{\num{3.28}} & \underline{\num{3.71}} & & \underline{\num{3.53}} \\
    \quad \phone + MPR & \num{12} & \num{768} & & \textbf{\num[text-series-to-math]{3.58}} & \underline{\num{4.49}} & & \textbf{\num[text-series-to-math]{1.79}} & \underline{\num{2.30}} & & \textbf{\num[text-series-to-math]{2.88}} & \textbf{\num[text-series-to-math]{3.35}} & & \textbf{\num[text-series-to-math]{3.07}} & & \underline{\num{3.92}} & \underline{\num{4.61}} & & \textbf{\num[text-series-to-math]{2.57}} & \textbf{\num[text-series-to-math]{3.08}} & & \textbf{\num[text-series-to-math]{2.86}} & \textbf{\num[text-series-to-math]{3.32}} & & \textbf{\num[text-series-to-math]{3.39}} \\
    \midrule
    \quad\textit{self-supervised FT (10 h)} \\
    HuBERT-base \\
    \quad + K-means (L\num{11}) & \num{10} & \num{768} & & \num{5.71} & \num{6.64} & & \num{3.15} & \num{4.09} & & \num{7.13} & \num{7.58} & & \num{5.72} & & \num{5.65} & \num{6.38} & & \num{4.79} & \num{5.40} & & \num{5.09} & \num{5.77} & & \num{5.51} \\
    \cdashline{1-25}
    \mphubert-\feat \\
    \quad + K-means (L\num{9}) & \num{10} & \num{768} & & \textbf{\num[text-series-to-math]{4.72}} & \textbf{\num[text-series-to-math]{5.50}} & & \underline{\num{2.58}} & \num{3.31} & & \num{5.08} & \num{5.59} & & \num{4.46} & & \num{5.01} & \textbf{\num[text-series-to-math]{5.56}} & & \num{4.19} & \underline{\num{4.71}} & & \num{4.38} & \num{5.00} & & \underline{\num{4.81}} \\
    \quad + K-means (feat) & \num{9} & \num{768} & & \num{5.00} & \num{5.81} & & \num{2.69} & \num{3.41} & & \num{5.29} & \num{5.69} & & \num{4.65} & & \num{5.16} & \num{5.92} & & \num{4.30} & \num{4.98} & & \num{4.51} & \num{5.20} & & \num{5.01} \\
    \quad + feat. freq. & \num{9} & \num{768} & & \num{4.88} & \underline{\num{5.65}} & & \num{2.63} & \num{3.28} & & \num{5.24} & \num{5.66} & & \num{4.56} & & \underline{\num{4.99}} & \num{5.80} & & \num{4.19} & \num{4.86} & & \num{4.39} & \num{5.09} & & \num{4.89} \\
    \quad + phone freq. & \num{9} & \num{768} & & \num{5.01} & \num{5.90} & & \num{2.62} & \num{3.35} & & \num{5.21} & \num{5.62} & & \num{4.62} & & \num{5.09} & \num{5.87} & & \num{4.31} & \num{5.01} & & \num{4.53} & \num{5.24} & & \num{5.01} \\
    \mphubert-\phone \\
    \quad + K-means (proj) & \num{10} & \num{768} & & \num{4.91} & \num{5.71} & & \num{2.66} & \num{3.32} & & \num{4.93} & \num{5.55} & & \num{4.51} & & \textbf{\num[text-series-to-math]{4.84}} & \underline{\num{5.62}} & & \num{4.17} & \num{4.81} & & \num{4.38} & \num{5.15} & & \num{4.83} \\
    \quad + K-means (phone) & \num{10} & \num{768} & & \num{4.88} & \num{5.83} & & \num{2.70} & \num{3.40} & & \num{5.29} & \num{5.79} & & \num{4.65} & & \num{5.52} & \num{6.16} & & \num{4.14} & \num{4.76} & & \num{4.28} & \underline{\num{4.86}} & & \num{4.95} \\
    \quad + phone freq. & \num{10} & \num{768} & & \underline{\num{4.77}} & \num{5.78} & & \textbf{\num[text-series-to-math]{2.49}} & \underline{\num{3.17}} & & \textbf{\num[text-series-to-math]{4.82}} & \textbf{\num[text-series-to-math]{5.26}} & & \textbf{\num[text-series-to-math]{4.38}} & & \num{5.11} & \num{5.79} & & \underline{\num{4.09}} & \num{4.72} & & \underline{\num{4.24}} & \underline{\num{4.86}} & & \textbf{\num[text-series-to-math]{4.80}} \\
    \quad + all phones & \num{10} & \num{768} & & \num{4.88} & \num{5.84} & & \textbf{\num[text-series-to-math]{2.49}} & \textbf{\num[text-series-to-math]{3.16}} & & \underline{\num{4.85}} & \underline{\num{5.28}} & & \underline{\num{4.42}} & & \num{5.15} & \num{5.89} & & \textbf{\num[text-series-to-math]{4.05}} & \textbf{\num[text-series-to-math]{4.70}} & & \textbf{\num[text-series-to-math]{4.20}} & \textbf{\num[text-series-to-math]{4.83}} & & \textbf{\num[text-series-to-math]{4.80}} \\
    \bottomrule
  \end{tabular}%
  }
  \caption{Acoustic discriminability scores (lower is better) over \num{5} development languages (sw, ta, th, tr, uk) and, as test languages, the \num{5} languages from the Zero Resource Speech Challenge \num{2017} (en, fr, zh, de, wo). The best layer for each model is selected based on the average ABX score on the development languages. The best scores are in \textbf{bold} and the second best are \underline{underlined}.}
  \label{tab:abx-scores}
  \vspace{-0.5em}
\end{table*}

\subsection{Training Details}

We train \mphubert variants for feature or phone recognition across the \num{55} languages drawn from the VoxCommunis Corpus.
Due to PanPhon's ternary feature representation, we exclude \mphubert-\feat predictions that correspond to zero-valued target features.
Furthermore, to handle \textit{multiphthongs} (hypernym of diphthong), we use a uniform heuristic so that the duration of the resulting \textit{monophthongs} is roughly the same.

The \feat and \phone variants are trained to minimise binary and multiclass cross-entropy losses, respectively, at the frame level
with the Adam optimiser \citep{kingma_15_adam}.
We use one V\num{100} GPU for \qty{40}{\kilo\relax} steps with a tri-stage learning rate schedule (\qty{4}{\kilo\relax} for warmup and \qty{16}{\kilo\relax} for decay) that peaks at \num{5e-5}.
Following \citet{conneau2020unsupervised}, we employ a language up-sampling strategy to balance the amount of data between low-resource and high-resource languages. (See \Cref{app:data} for more details.)

\subsection{Evaluation and Results}

We evaluate our \mphubert models using three speech recognition metrics: frame-wise feature accuracy, frame-wise phone accuracy and phone error rate (PER) without deduplication heuristics.
For feature accuracy, we compute scores over non-zero features only, excluding zero-valued target features as in training.
Since \mphubert-\phone lacks an explicit feature space, we extract feature vectors from predicted phones using PanPhon's feature table.

\Cref{tab:feat-phone-acc} shows results for both \mphubert variants on held-out test sets from the \num{55} training languages and the \num{5} development languages.
Both variants exhibit superior performance on training languages, particularly for phone-level metrics.
When transitioning from training to development languages, phone accuracy drops by about \num{15} pp and PER increases by approximately \num{20} pp, while feature accuracy shows more resilience with only \numrange[ range-phrase = --, range-units = single]{3}{4} pp degradation.
The \feat variant consistently outperforms the \phone variant in articulatory feature prediction across language sets and features (see \Cref{fig:radar_fr}).
However, this advantage does not translate to improved phone recognition performance.
The \phone variant leads by \num{0.4} pp in phone accuracy and, since frame-level errors accumulate over the sequence, the gap widens to about \num{2} pp in PER.
Note that both gaps are stable across language sets.

\section{Few-shot Language Adaptation}
\label{sec:few-shot-adapt}

In this section, we assess the linguistic relevance of \mphubert's learnt representations by evaluating their phonetic invariance across languages and speaking styles, in a zero-shot or few-shot setting.

\subsection{Language Adaptation Setting}

\paragraph{Modes.}
We compare how SSL models encode speech in a new language in three modes: zero-shot, supervised fine-tuning and self-supervised fine-tuning.
All the baselines and our two \mphubert models are evaluated in zero-shot mode, while only the monolingual baseline and our two models are evaluated in the fine-tuning modes (on the \qty{10}{\hour} training split) for fairness\footnote{HuBERT-base and our \mphubert models are trained on two to three orders of magnitude less data than the multilingual baselines.}.
In the supervised mode, the models are trained to predict the ground-truth phones of masked inputs (MPR) or without masking at all (PR).
In the self-supervised mode, a clustering step first produces discrete pseudo-labels, which are later used as targets for masked prediction.

\paragraph{Baselines.}
We compare \mphubert against several baselines, including traditional acoustic features (MFCCs), the monolingual HuBERT-base backbone, and three self-supervised models trained on massively multilingual data: MMS-\num{1}B \citep{pratap24_mms}, mHuBERT-\num{147} \citep{zanonboito24_mhubert}, and XEUS \citep{chen24_xeus}.
\Cref{tab:baselines} shows a brief comparison of the training data between the baselines and our models.

\begin{figure}[t]
    \centering
    \begin{tikzpicture}
        \node (img) [anchor=south west, inner sep=0pt] {\includegraphics[width=0.47\textwidth]{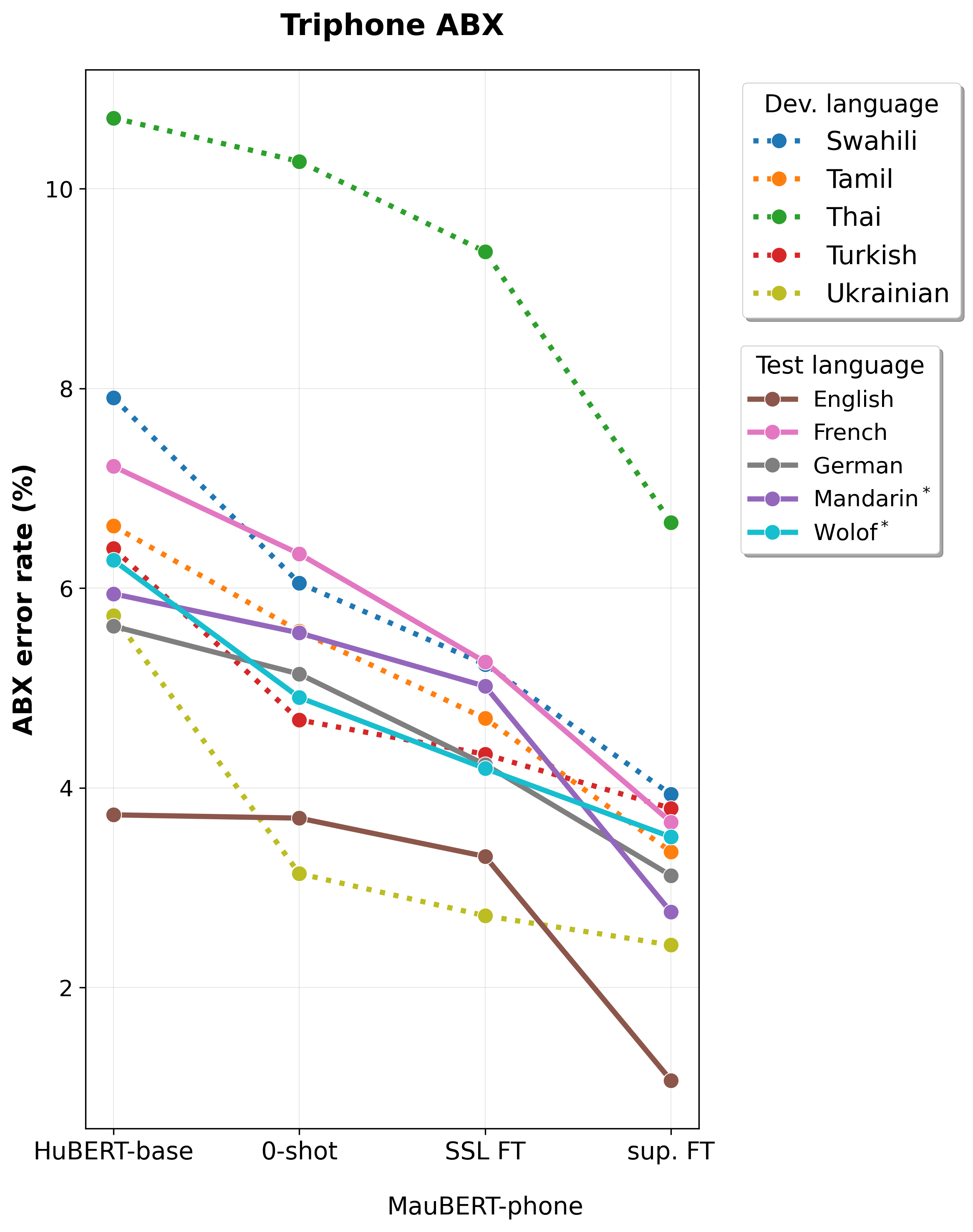}};
        \draw (2.0,0.43) -- (2.0,0.55);
        \draw (2.0,0.43) -- (5.55,0.43);
        \draw (5.55,0.43) -- (5.55,0.55);
    \end{tikzpicture}
    \caption{Reduction of the triphone ABX error rates across the \num{5} development languages and \num{5} test languages between the base HuBERT model, and \mphubert, tested in zero-shot and after masked fine-tuning (\qty{10}{\hour}) with or without labels on a new language. The two speaker conditions are averaged, and the \qty{10}{\second} subset is chosen for the test languages. $^*$Mandarin and Wolof only have \qtylist[list-units = single]{1.5;1.8}{\hour} of training data, resp.}
    \label{fig:triphone_abx_evol}
\end{figure}

\paragraph{Implementation.}
For the supervised fine-tuning, we train the HuBERT modules\footnote{Unlike pre-training, the feature extractor is also updated.} for \qty{20}{\kilo\relax} steps on one V\num{100} GPU with a tri-stage learning rate schedule (\qty{2}{\kilo\relax} for warmup and \qty{8}{\kilo\relax} for decay).
We use the Adam optimiser with a peak learning rate at \num{1e-4}.
For the self-supervised fine-tuning, we train these modules for \qty{50}{\kilo\relax} steps on one H\num{100} GPU.
We use the Adam optimiser with a linear decay schedule (\qty{8}{\percent} for warmup, then linear decay back to zero) that peaks at \num{5e-6}.

\subsection{Metric}
We employ the ABX discriminability test to measure phonetic invariance \citep{schatz16_phdthesis}.
It evaluates speech representations by comparing distances between three triphones: \textit{A}, \textit{X} (same linguistic unit as \textit{A}), and \textit{B} (different unit).
The test is considered successful when the distance between \textit{A} and \textit{X} is smaller than that between \textit{A} and \textit{B}.
The test comprises two variants: a triphone-based version that examines complete triphone representations, and a phoneme-based version that focuses exclusively on central phone representations.

The speaker condition varies between two scenarios: \textit{within}-speaker (all triphones share the speaker) and \textit{across}-speaker (only \textit{A} and \textit{B} share the speaker).
In addition, contextual conditions across all three items (\textit{A}, \textit{B}, and \textit{X}) can be manipulated: \textit{within}-context (where all items share identical surrounding phonetic context) versus \textit{any}-context (where surrounding contexts may differ).

We compute all the ABX scores with the CPU backend of \texttt{fastabx} \citep{poli25_fastabx}.

\subsection{Language Data}
\label{sec:abx_data}

Following the Zero Resource Speech Challenge \num{2017} \cite{dunbar17_zerospeech}, we curate ABX-ready datasets for five \textit{development languages} from the VoxCommunis Corpus: Swahili, Tamil, Thai, Turkish and Ukrainian.
The ABX datasets consist of three splits for each language: a \qty{10}{\hour} training set, a validation set and a test set.
We select the best parameters, hyperparameters and layers of the various models according to their impact on the average ABX score (triphone-based ABX, within-context phoneme ABX and any-context phoneme ABX) on the ABX test sets.

We use both the development and surprise languages from the aforementioned Zero Resource Speech Challenge \num{2017}, namely English, French, Mandarin, German and Wolof, as \textit{test languages} (hereafter referred to as ZRC\num{2017}).
The amount of speech in the original training set ranges from \qty{2.3}{\hour} for Mandarin to \qty{35.3}{\hour} for English.
We thus extract training and validation splits of up to \qty{10}{\hour}.
In line with the desiderata of the challenge, we keep the original test subsets of differing length (\qty{1}{\second}, \qty{10}{\second} and \qty{120}{\second}) to evaluate the effect of context length (triphone-based ABX only).
We evaluate only the best configuration for each model on these languages.

Additionally, we curate an ABX dataset of casual speech in English and French, sourcing high-quality recorded conversations of native speakers.
The dataset possesses the same three-split structure as the development languages.

\begin{table}[t]
  \centering
  \resizebox{0.47\textwidth}{!}{%
      \begin{tabular}{ l c c c c c }
        \toprule
        \multirow{2}{*}{Systems} & \multicolumn{2}{c}{Read} & & \multicolumn{2}{c}{Casual} \\
        \cline{2-3} \cline{5-6}
        & WS & AS & & WS & AS \\
        \midrule
        MMS & \num{6.75} & \num{8.99} & & \num{13.47} & \num{17.47} \\
        XEUS & \num{4.46} & \num{5.78} & & \num{8.69} & \num{11.29} \\
        mHuBERT-147 & \num{5.29} & \num{6.83} & & \num{10.62} & \num{13.72} \\
        HuBERT-base & \num{4.71} & \num{6.24} & & \num{9.41} & \num{12.48} \\
        \cdashline{1-6}
        \quad \textit{ours} \\
        \mphubert-\feat & \num{4.45} & \num{5.75} & & \num{9.43} & \num{12.11} \\
        \quad + K-means (L\num{9}) & \underline{\num{3.95}} & \underline{\num{5.00}} & & \textbf{\num[text-series-to-math]{8.25}} & \textbf{\num[text-series-to-math]{10.66}} \\
        \mphubert-\phone & \num{4.29} & \num{5.75} & & \num{9.23} & \num{11.92} \\
        \quad + phone freq. & \textbf{\num[text-series-to-math]{3.69}} & \textbf{\num[text-series-to-math]{4.89}} & & \underline{\num{8.43}} & \underline{\num{10.85}} \\
        \bottomrule
      \end{tabular}%
  }
  \caption{Triphone-based ABX error rates across registers (read vs. spontaneous) for English and French in zero-shot mode. Our two \mphubert variants are also tested after self-supervised fine-tuning on \qty{10}{\hour}.}
  \label{tab:casual-abx}
\end{table}

\subsection{Few-shot Language Adaptation Results}
\label{sec:few-shot-adapt-results}

Our experimental results demonstrate the competitive performance of our \mphubert models across multiple evaluation scenarios.

\paragraph{Multilingual Training Benefits.}
The top of \Cref{tab:abx-scores} illustrates the phonetic invariance performance in zero-shot mode achieved by \mphubert models through multilingual pre-training.
Our models attain particularly strong results in the any-context phoneme-based ABX tasks, and the \mphubert-\phone model delivers the best overall zero-shot performance (\qty{5.22}{\percent} against \qty{5.74}{\percent} for XEUS).
Further, \Cref{fig:triphone_abx_evol} confirms the performance improvements of the multilingual pre-training as well as the proposed self-supervised fine-tuning: \qty{6.62}{\percent} for the HuBERT-base baseline vs.~\qty{5.54}{\percent} for \mphubert-\phone in zero-shot and \qty{4.84}{\percent} after \textit{phone freq.}~MPR in triphone ABX with similar audio lengths.

\paragraph{Cross-linguistic Performance Patterns.}
\Cref{tab:abx-scores} also shows that development languages present greater challenges than test languages when evaluated under comparable conditions (triphone ABX with similar audio lengths) across models and modes.
This pattern indicates varying degrees of phonetic complexity across language families and suggests that our model selection strategy (detailed in \S \ref{sec:abx_data}) based on development languages' performance provides a robust foundation for cross-lingual generalisation.
\Cref{fig:triphone_abx_evol} reinforces this observation, with development languages (shown with dotted lines) generally exhibiting higher error rates and more variable performance across the training progression compared to test languages (solid lines), suggesting the former present more challenging phonetic discrimination tasks and may represent more diverse or complex phonological systems.

\paragraph{Supervised Fine-tuning Efficacy.}
Supervised fine-tuning yields substantial improvements in ABX error rates.
Particularly striking is the effectiveness of predicting the ground-truth phones of masked inputs (MPR), which reduces ABX error rates compared to standard phone prediction (PR), especially for triphone-based ABX.
The \mphubert-\phone + MPR configuration achieves the best supervised performance (\qty{3.07}{\percent} on development languages, \qty{3.39}{\percent} on test languages), representing a significant \qty{38}{\percent} relative improvement over the zero-shot baseline. \Cref{fig:triphone_abx_evol} illustrates this systematic improvement pattern across all languages, with supervised fine-tuning showing the most notable gains (\qty{3.43}{\percent} average ABX score).
Remarkably, fine-tuning effectiveness appears largely independent of training data quantity: low-resource language Wolof achieves comparable error rates to high-resource languages, indicating robust few-shot adaptation capabilities.

\paragraph{Self-supervised Fine-tuning Analysis.}
While self-supervised fine-tuning approaches show consistent improvements over zero-shot performance, a performance gap remains compared to the fully-supervised standard.
Among the clustering strategies, our phone frequency-based approach demonstrates some gains over standard K-means clustering, particularly excelling in phoneme-level discrimination tasks and longer temporal contexts (\qty{10}{\second} and \qty{120}{\second} triphone ABX).
\mphubert-\phone with phone frequency clustering achieves the best self-supervised performance (\qty{4.59}{\percent} average ABX score), highlighting the value of linguistically-informed clustering strategies.

\paragraph{Speech Register Adaptation Results.}
\Cref{tab:casual-abx} reveals nuanced domain-specific patterns across read versus casual speech.
In zero-shot mode, our models perform slightly better than multilingual baselines on read speech (\mphubert-\phone: \qty{5.02}{\percent} vs. XEUS: \qty{5.12}{\percent}) but show reversed performance on casual speech (\qty{10.58}{\percent} vs.~\qty{9.99}{\percent} for XEUS), reflecting the inherent difficulty of spontaneous speech processing with its increased phonetic variability and reduced articulatory precision.
However, self-supervised fine-tuning not only amplifies our advantage on read speech (\qty{4.29}{\percent}) but also recovers competitive performance on casual speech (\qty{9.64}{\percent}), demonstrating the robustness of our adaptation approach across speech domains.

\subsection{Phonetic Inventory Discovery Results}

Two of the \mphubert SSL methods consist in assigning a feature or a phoneme set to a new language as a target for SSL fine-tuning.
These methods amount to discovering the  \textit{phonetic} inventories of previously unseen languages\footnote{\mphubert-\feat can only predict monophthongs due to the splitting of \textit{multiphthongs} during training.}.
Following \citet{zelasko22_asrinventories}, we leverage the frequency distribution of (discrete) articulatory feature vectors produced by \mphubert-\feat, where high-frequency combinations likely correspond to actual phones in the language inventory\footnote{The inventory consists of all the phones observed in VoxCommunis. Most phones appear in the `CV dictionaries' on \url{https://mfa-models.readthedocs.io/en/latest/dictionary/index.html}.}.

\Cref{tab:inventory_discovery} reveals a clear trade-off between precision and recall across different threshold strategies. The top-\num{100} approach achieves consistently high recall (at least \num{0.825} for four out of five languages), successfully capturing most phonemes in the target inventories.
However, this comes at the cost of precision (\numrange[range-phrase = --]{0.270}{0.390}), indicating substantial inclusion of spurious feature vectors.
Conversely, the optimised frequency threshold approach significantly improves precision (\numrange[range-phrase = --]{0.778}{0.872}) while maintaining reasonable recall (\numrange[range-phrase = --]{0.532}{0.810}), suggesting more accurate phonetic identification with fewer false positives.

The superior F\textsubscript{\num{1}} performance of optimised thresholds over fixed thresholds underscores the importance of adaptive, data-driven approaches to inventory discovery.
(See \Cref{tab:inventory_examples} for some inventory examples with F\textsubscript{\num{1}}-optimal thresholds.)

\section{Discussion}

\paragraph{Broader Impact.}
Our demonstration that effective phonetic models can be developed for low-resource languages with minimal training data (as evidenced by Wolof performance with less than \qty{2}{\hour} of data) is an encouraging signal towards more linguistic inclusion in computational models.
In addition, our frequency-based methodology offers particular value for endangered language documentation, where traditional phonological analysis may be impractical, providing linguists with not only a multilingual articulatory feature recogniser but also an automated tool for initial phonetic hypothesis generation that can guide subsequent detailed analysis.
However, the superior performance of high-resource languages like English also highlights the importance of linguistic diversity in training data, since the imbalance thereof could persist through evaluation.

\paragraph{Future Work.}
Several promising research directions emerge from our findings.
The counter-intuitive relationship between training data quantity and fine-tuning effectiveness suggests that investigation into optimal data selection strategies could yield significant improvements, potentially focusing on phonetically diverse rather than simply large datasets.
Given that speaker and content information are encoded at different layers of HuBERT's encoder, selectively targeting a subset of hidden layers during multilingual pre-training is another promising avenue worth exploring.
The domain adaptation capabilities demonstrated in our casual speech experiments indicate potential for developing more robust models through multi-domain training paradigms.
Furthermore, extending the self-supervised fine-tuning beyond the encoder to encompass the entire \mphubert architecture could address current limitations by enabling end-to-end adaptation of both the pre-trained representations and the downstream articulatory feature prediction modules, potentially leading to improved performance on target languages and domains, and better phonetic inventory discovery.

\section{Conclusion}
\label{sec:conclusion}

This work presents \mphubert, a multilingual extension of HuBERT that demonstrates competitive phonetic discrimination capabilities across diverse languages while revealing important insights about cross-lingual representation learning.
Our results establish that multilingual supervised pre-training creates robust phonetic foundations that enable effective few-shot adaptation to new languages (\num{10} hours of speech) with or without supervision.
The demonstrated effectiveness on both read and spontaneous speech, coupled with strong performance on low-resource languages, positions this work as a significant step towards more inclusive multilingual speech technologies.

\section*{Limitations}

The evaluation is constrained to the ABX discrimination task, which, while established as a standard phonetic benchmark, may not fully capture the nuanced linguistic representations as required for other linguistic levels (\textit{e.g.} syntax and semantics).
The performance gap between self-supervised and supervised fine-tuning methods suggests that clustering- or frequency-based approaches, despite their linguistic motivation, remain suboptimal compared to gold-standard supervision.

\ifaclpubformat
\section*{Acknowledgments}

We thank the reviewers for their valuable feedback, which helped improve the clarity and quality of the paper.
This work was granted access to the HPC resources of IDRIS under the allocation 2024-AD011014739R1 made by GENCI, and was supported in part by Agence Nationale de Recherche (ANR-17-EURE-0017 FrontCog, ANR-10-IDEX-0001-02 PSL and ANR-23-IACL-0006 France 2030).
ED in his EHESS role and MK were funded by an ERC grant (InfantSimulator). Views and opinions expressed are those of the authors only and do not necessarily reflect those of the European Union or the European Research Council.
Neither the European Union nor the granting authority can be held responsible for them.
\fi

\newpage

\bibliography{custom}

\appendix

\appendix

\newpage
\begin{table*}[t]
    \centering
    \resizebox{\textwidth}{!}{%
    \addtolength{\tabcolsep}{-0.1em}
        \begin{tabular}[t]{c|c|c}
            \toprule
            IETF code & Language & \# Hours \\
            \midrule
            ab & Abkhaz & \num{22.4} \\
            am & Amharic & \num{0.1} \\
            ba & Bashkir & \num{49.7} \\
            be & Belarusian & \num{50.0} \\
            bg & Bulgarian & \num{6.2} \\
            bn & Bengali & \num{30.5} \\
            ca & Catalan & \num{50.0} \\
            ckb & Central kurdish & \num{6.6} \\
            cs & Czech & \num{24.9} \\
            cv & Chuvash & \num{0.5} \\
            dv & Maldivian & \num{2.5} \\
            el & Greek & \num{2.1} \\
            eu & Basque & \num{50.0} \\
            gn & Guarani & \num{1.5} \\
            ha & Hausa & \num{2.2} \\
            hi & Hindi & \num{4.6} \\
            hsb & Upper sorbian & \num{1.5} \\
            hu & Hungarian & \num{49.4} \\
            hy-AM & Armenian & \num{0.4} \\
            \bottomrule
        \end{tabular}
        \quad
        \begin{tabular}[t]{c|c|c}
            \toprule
            IETF code & Language & \# Hours \\
            \midrule
            id & Indonesian & \num{7.4} \\
            it & Italian & \num{50.0} \\
            ja & Japanese & \num{12.1} \\
            ka & Georgian & \num{50.0} \\
            kk & Kazakh & \num{0.0} \\
            kmr & Northern kurdish & \num{4.8} \\
            ko & Korean & \num{0.6} \\
            ky & Kyrgyz & \num{2.2} \\
            lij & Ligurian & \num{0.7} \\
            lt & Lithuanian & \num{9.4} \\
            ml & Malayalam & \num{1.4} \\
            mn & Mongolian & \num{3.1} \\
            mr & Marathi & \num{3.6} \\
            mt & Maltese & \num{2.2} \\
            myv & Erzya & \num{1.9} \\
            nan-tw & Taiwanese hokkien & \num{2.0} \\
            pa-IN & Punjabi & \num{1.1} \\
            nl & Dutch & \num{40.4} \\
            \bottomrule
        \end{tabular}
        \quad
        \begin{tabular}[t]{c|c|c}
            \toprule
            IETF code & Language & \# Hours \\
            \midrule
            pl & Polish & \num{28.9} \\
            pt & Portuguese & \num{23.8} \\
            ro & Romanian & \num{3.9} \\
            ru & Russian & \num{37.3} \\
            rw & Kinyarwanda & \num{50.0} \\
            sk & Slovak & \num{3.3} \\
            sl & Slovenian & \num{1.3} \\
            sq & Albanian & \num{0.1} \\
            sr & Serbian & \num{1.4} \\
            sv-SE & Swedish & \num{8.2} \\
            tk & Turkmen & \num{1.1} \\
            tt & Tatar & \num{9.3} \\
            ug & Uyghur & \num{15.2} \\
            ur & Urdu & \num{0.1} \\
            uz & Uzbek & \num{50.0} \\
            vi & Vietnamese & \num{1.4} \\
            yo & Yoruba & \num{1.9} \\
            yue & Cantonese & \num{3.3} \\
            \midrule
            & Total & \num{788.4} \\
            \bottomrule
        \end{tabular}%
    }
    \caption{List of \num{55} languages with their amount of speech included in the training set.}
    \label{tab:data}
\end{table*}

\section{Data}
\label{app:data}

We use the annotations of Common Voice \citep{ardila20_common_voice} made by \citet{ahn22_voxcommunis}.
At the time of download (21 August 2024), the dataset consisted of \num{63} languages.
Five languages (Swahili, Tamil, Thai, Turkish and Ukrainian) were held out for hyperparameter tuning, and three languages (French, Mandarin and Hong Kong Mandarin) were discarded because of their presence in the test set.
\Cref{tab:data} contains the list of the \num{55} languages kept for training.

\paragraph{Training languages.}
We filter out utterances containing \texttt{spn} segments, which indicate alignment errors from Montreal Forced Aligner \citep{mcauliffe17_mfa}, or those that are excessively long (non-silent phones exceeding \qty{0.5}{\second}).
We retain only utterances lasting between \qtylist[list-units = single]{2;20}{\second}.
We fix misplaced diacritics that were incorrectly attached to adjacent phones in four languages.
We also handle IPA characters that PanPhon \citep{mortensen16_panphon} does not recognise by mapping them to their proper equivalents (\textit{e.g.} \textipa{[\textg]} becomes \textipa{[g]}).
To prevent high-resource languages from dominating the training data for \mphubert, we limit each language to a maximum of \qty{50}{\hour}, yielding a total of \qty{788.4}{\hour} of multilingual training speech.

\paragraph{Development languages.}
We apply the same preprocessing pipeline used for the training languages to the development languages.
Next, we combine the three original Common Voice splits (train, dev, and test) and create three new splits with the following specifications for each language:
\begin{enumerate*}[label=(\roman*)]
    \item the test set contains \qtyrange[range-phrase = --, range-units = single]{7.3}{14.0}{\hour} of audio uniformly distributed across \num{20} speakers,
    \item the training set contains \qtyrange[range-phrase = --, range-units = single]{8.3}{9.5}{\hour} uniformly distributed across \num{10} speakers, and
    \item the validation set contains \qtyrange[range-phrase = --, range-units = single]{8.5}{9.7}{\hour} hours of audio.
\end{enumerate*}
We ensure that speakers are completely disjoint across all newly created splits.

\paragraph{Test languages.}
We use the five languages from the Zero Resource Challenge \num{2017} \cite{dunbar17_zerospeech}: English, French, Mandarin, German, and Wolof.
From the original long-form recordings (each corresponding to a different speaker), we extract\footnote{The challenge training set contains at least \qty{21}{\hour} of speech, except for Mandarin and Wolof, which have \qtylist[list-units = single]{2.3;3.0}{\hour} of speech, respectively.} training and validation splits of up to \qty{10}{\hour} per language through the following process:
\begin{enumerate*}[label=(\roman*)]
    \item apply voice activity detection using official challenge alignments,
    \item segment recordings into \qtyrange[range-phrase = --, range-units = single]{2}{20}{\second} clips including silences of up to \qty{1}{\second}, and
    \item assign each speaker's clips exclusively to either training or validation splits to ensure speaker disjointness.
\end{enumerate*}
This yields training sets of \qtyrange[range-phrase = --, range-units = single]{1.5}{10.0}{\hour} and validation sets of \qtyrange[range-phrase = --, range-units = single]{0.7}{10.0}{\hour}, with Mandarin having the smallest splits and European languages having the largest.

\section{Training}
\label{app:hyperparam}
\Cref{tab:hyper_param} lists the (hyper-) parameters used for multilingual feature recognition.
All the (hyper-) parameters for self-supervised and supervised fine-tuning can be found in the released code.

\paragraph{Language Up-sampling.}
During multilingual pre-training, we draw from the multinomial distribution $p_l \sim (\frac{n_l}{N})^{\alpha}$, where $n_l$ is the number of audios of language $l$, $N$ is the training set size, and $\alpha$ is the up-sampling factor controlling the importance between high- and low-resource languages.

\paragraph{Length grouping.}
To reduce unused representations in batches, we split the multilingual data into buckets of audio of roughly the same length.

\section{Clustering methods}
\label{app:clustering}

\paragraph{Ground-truth phones.}
We use the collapsed list of segments from PanPhon for the development languages, and the list of unique phonemes from the official alignments for the test languages.

\paragraph{K-means.}
We run the \texttt{MiniBatchKMeans} algorithm from \texttt{scikit-learn} \citep{pedregosa_11_scikit} on the training set for each development and test language.
We select three different representations:
\begin{enumerate*}[label=(\roman*)]
    \item the best-performing layer from zero-shot mode,
    \item the feature logits for \mphubert-\feat,
    \item and the phone logits (after reducing to only the phones seen during training) for \mphubert-phone.
\end{enumerate*}

\paragraph{Predicted phones.}
First, we determine the most likely phone label for each frame in the training set. We then prune the phone prediction layer by removing the output heads corresponding to phones absent from the training set (\textit{all phones} labels). For the \textit{phone freq.} labels, we further restrict the prediction layer to the $K$ most frequent phones in the training set, discarding the remaining heads and fine-tuning the layer to produce highly confident predictions for these $K$ phones.

\paragraph{Predicted features.}
For the \textit{feat. freq.} labels, we extract predicted articulatory feature logits for each frame in the training set, apply a sigmoid activation, and hard-threshold the resulting values at \num{0.5} to obtain binary feature vectors. We then compute the frequency of each unique binary vector across the training set and retain the $K$ most common ones. For pseudo-labelling, each frame is assigned to its nearest neighbour among these $K$ frequent vectors under the $\ell_1$ distance, with ties broken in favour of vectors with fewer zero-valued features.

\begin{table*}[t]
    \centering
    \resizebox{0.92\textwidth}{!}{%
    \begin{tabular}[t]{ l c }
        \toprule
        Paramenters & Value \\
        \midrule
        \textbf{Model} \\
        \quad Up-projection dimension & \num{1024} \\
        \quad BLSTM layers & \num{2} \\
        \quad BLSTM dimension & \num{1024} \\
        \quad BLSTM dropout & \num{0.2} \\
        \quad BLSTM layer normalisation & No \\
        \quad Phone MLP hidden dimension & \num{1024} \\
        \quad Phone MLP activation function & GELU \\
        \textbf{Features} \\
        \quad Diphthong feature strategy & split \\
        \quad Zero values loss & ignore \\
        \bottomrule
    \end{tabular}
    \qquad\qquad\qquad
        \begin{tabular}[t]{ l c }
        \toprule
        Hyper-Paramenters & Value \\
        \midrule
        \textbf{Data} \\
        \quad Up-sample factor ($\alpha$) & \num{0.7} \\
        \quad Batch size & \num{32} \\
        \textbf{Optimizer} \\
        \quad Name & Adam \\
        \quad Peak learning rate & \num{5e-5} \\
        \quad Betas & (\num{0.9}, \num{0.98}) \\
        \quad Weight decay & No \\
        \quad Epsilon & \num{1e-8} \\
        \quad Warmup steps & \num{4000} \\
        \quad Hold steps & \num{16000} \\
        \quad Decay steps & \num{20000} \\
        \quad Mixed precision & fp16 \\
        \bottomrule
    \end{tabular}%
    }
    \caption{Model parameters and training hyper-parameters used for \mphubert-\feat.}
    \label{tab:hyper_param}
\end{table*}

\section{Inventory discovery supplement}

\begin{table}[H]
  \centering
  \resizebox{0.47\textwidth}{!}{%
      \addtolength{\tabcolsep}{-0.3em}
      \begin{tabular}{ l c c c c c c c c }
        \toprule
        \multirow{2}{*}{Language} & \multirow{2}{2cm}{\centering Inventory size} & \multicolumn{3}{c}{Top \num{100}} & & \multicolumn{3}{c}{F\textsubscript{1}-optimal} \\
        \cline{3-5} \cline{7-9}
        & & Prec. & Recall & F\textsubscript{1} & & Prec. & Recall & F\textsubscript{1} \\
        \midrule
        Swahili & \num{40} & \num{0.330} & \num{0.825} & \num{0.471} & & \num{0.824} & \num{0.700} & \num{0.757} \\
        Tamil & \num{35} & \num{0.300} & \num{0.857} & \num{0.444} & & \num{0.815} & \num{0.629} & \num{0.710} \\
        Thai & \num{42} & \num{0.390} & \num{0.929} & \num{0.549} & & \num{0.872} & \num{0.810} & \num{0.840} \\
        Turkish & \num{47} & \num{0.270} & \num{0.574} & \num{0.367} & & \num{0.781} & \num{0.532} & \num{0.633} \\
        Ukrainian & \num{38} & \num{0.350} & \num{0.921} & \num{0.507} & & \num{0.778} & \num{0.737} & \num{0.757} \\
        \bottomrule
      \end{tabular}%
  }
  \caption{Precision, recall and F\textsubscript{\num{1}} score for the inventory discovery on the development languages for the top-\num{100} threshold and the best threshold for F\textsubscript{\num{1}} score.}
  \label{tab:inventory_discovery}
\end{table}

\begin{table}[h]
  \centering
  \resizebox{0.47\textwidth}{!}{%
      \addtolength{\tabcolsep}{-0.2em}
      \begin{tabular}{ l c c }
        \toprule
        Lang. & Correctly predicted phones & Missing phones \\
        \midrule
        \multirow{3}{*}{\centering th} & \multirow{3}{4.5cm}{\centering \textipa{m}, \textipa{i}, \textipa{k}, \textipa{j}, \textipa{u}, \textipa{a}, \textipa{p}, \textipa{w}, \textipa{n}, \textipa{t}, \textipa{l}, \textipa{s}, \textipa{b}, \textipa{N}, \textipa{e}, \textipa{o}, \textipa{h}, \textipa{d}, \textipa{f}, \textipa{E}, \textipa{O}, \textipa{i:}, \textipa{a:}, \textipa{u:}, \textipa{R}, \textipa{e:}, \textipa{k\super h}, \textipa{p\super h}, \textipa{t\super h}, \textipa{E:}, \textipa{O:}, \textipa{\t{tC}}, \textipa{7}, \textipa{0:}} & \multirow{3}{3.4cm}{\centering \textipa{P}, \textipa{o:}, \textipa{W}, \textipa{\t{tC}\super h}, \textipa{W:}, \textipa{7:}, \textipa{0}, \textipa{\textsubarch{a}}} \\
        \\ \\
        \cdashline{1-3}
        \multirow{3}{*}{\centering tr} & \multirow{3}{4.5cm}{\centering \textipa{m}, \textipa{i}, \textipa{k}, \textipa{j}, \textipa{u}, \textipa{a}, \textipa{p}, \textipa{b}, \textipa{e}, \textipa{o}, \textipa{g}, \textipa{h}, \textipa{f}, \textipa{\t{tS}}, \textipa{S}, \textipa{\t{dZ}}, \textipa{R}, \textipa{t}, \textipa{n}, \textipa{d}, \textipa{W}, \textipa{y}, \textipa{s}, \textipa{l}, \textipa{z}} & \multirow{3}{3.4cm}{\centering \textipa{m:}, \textipa{k:}, \textipa{j:}, \textipa{p:}, \textipa{b:}, \textipa{g:}, \textipa{h:}, \textipa{f:}, \textipa{\t{tS}:}, \textipa{S:}, \textipa{\t{dZ}:}, \textipa{v}, \textipa{v:}, \textipa{R:}, \textipa{t:}, \textipa{n:}, \textipa{Z}, \textipa{d:}, \textipa{s:}, \textipa{\oe}, \textipa{l:}, \textipa{z:}} \\
        \\ \\
        \cdashline{1-3}
        \multirow{3}{*}{\centering uk} & \multirow{3}{4.5cm}{\centering \textipa{m}, \textipa{i}, \textipa{k}, \textipa{j}, \textipa{u}, \textipa{p}, \textipa{b}, \textipa{r}, \textipa{E}, \textipa{S}, \textipa{t}, \textipa{x}, \textipa{\t{tS}}, \textipa{n}, \textipa{Z}, \textipa{d}, \textipa{a}, \textipa{s}, \textipa{t\super j}, \textipa{l\super j}, \textipa{v}, \textipa{z}, \textipa{\t{ts}}, \textipa{s\super j}, \textipa{r\super j}, \textipa{l}, \textipa{n\super j}, \textipa{\t{ts}\super j}} & \multirow{3}{3.4cm}{\centering \textipa{g}, \textipa{f}, \textipa{O}, \textipa{\t{dZ}}, \textipa{I}, \textipa{H}, \textipa{\t{dz}}, \textipa{d\super j}, \textipa{z\super j}, \textipa{\t{dz}\super j}} \\
        \\ \\
        \bottomrule
      \end{tabular}%
  }
  \caption{Phonetic inventory prediction using an F\textsubscript{1}-optimal threshold for Thai, Turkish and Ukrainian. The language inventories comprise all the phones observed in the alignments from VoxCommunis.}
  \label{tab:inventory_examples}
\end{table}

\section{Spoken language modelling evaluation}
\label{app:spoken-lm}

To complement the ABX phonetic probing results, we assess whether the discrete units produced by \mphubert lead to better spoken language models (sLMs) through using these units as input features to a spoken LM trained in a larger dataset (\qty{6}{\kilo\relax} hours of Libri-Light). 

We use our zero-shot models and SSL models fine-tuned on the English adaptation set (\qty{10}{\hour}) as frame-level encoders.
A K-means model (with $V$ clusters) is trained on the resulting embeddings to produce discrete tokens, which are then used to train an OPT-\num{125}M language model \citep{zhang2022optopenpretrainedtransformer} with fairseq2 \citep{balioglu2023fairseq2}, following the architectural choices of \citet{hassid2023textually} and \citet{poli2025spidr}.
Training uses the \qty{6}{\kilo\relax} hours subset of Libri-Light \citep{9052942} on \num{8} GPUs, with a context length of \num{2048} tokens, batches of at most \num{81920} tokens, for \qty{25}{\kilo\relax} steps.
The learning rate is set to \num{1e-2} with a linear warmup of \num{1000} steps and cosine annealing; remaining hyper-parameters follow OPT-\num{125}M defaults.
We choose the checkpoint with the lowest validation loss.

We compare \mphubert against the monolingual SSL models: wav2vec 2.0 \citep{baevski2020wav2vec}, HuBERT-base \citep{hsu21_hubert}, WavLM-base \cite{chen_22_wavlm} and DinoSR \cite{liu2023dinosr}, all trained on \qty{1}{\kilo\relax} hours of English.
We evaluate sLMs on three tasks: sWUGGY (lexical), sBLIMP (syntactic), and tSC (semantic consistency), reporting scores as percentages.
We consider two vocabulary sizes: $V=256$ (\Cref{tab:slm-scores}) and $V=100$ (\Cref{tab:slm-scores-100}).
For \Cref{tab:slm-scores}, the \mphubert-\phone zero-shot (all phones) variant uses a one-hot encoding over all \num{370} observed phones in all our data; for \Cref{tab:slm-scores-100}, the \mphubert-\phone SSL fine-tuned (logits) variant uses a one-hot encoding over all \num{87} observed phones in the English adaptation data.

\begin{table}[t]
    \centering
    \resizebox{0.47\textwidth}{!}{%
        \addtolength{\tabcolsep}{-0.2em}
        \begin{tabular}{lccccc}
            \toprule
            \multirow{2}{*}{Model} & \multirow{2}{*}{Units} & \multicolumn{2}{c}{sWUGGY} & \multirow{2}{*}{sBLIMP} & \multirow{2}{*}{tSC} \\
            \cmidrule(lr){3-4}
             & & all & in-vocab & & \\
            \midrule
            \quad\textbf{\textit{Monolingual models}} \\
            wav2vec 2.0$^\dagger$ & K-means (L\num{6}) & \num{62.29} & \num{68.50} & \num{53.34} & \num{65.97} \\
            WavLM-base & K-means (L\num{11}) & \textbf{\num[text-series-to-math]{70.07}} & \textbf{\num[text-series-to-math]{80.36}} & \num{56.48} & \textbf{\num[text-series-to-math]{71.15}} \\
            HuBERT-base & K-means (L\num{11}) & \num{65.01} & \num{72.96} & \num{55.34} & \num{68.22} \\
            DinoSR$^\dagger$ & codebook (L\num{5}) & \num{60.10} & \num{64.56} & \num{57.04} & \num{69.44} \\
            SpidR$^\dagger$ & codebook (L\num{6}) & \underline{\num{69.78}} & \underline{\num{79.98}} & \textbf{\num[text-series-to-math]{58.10}} & \underline{\num{70.14}} \\
            \addlinespace
            \quad\textbf{\textit{Ours}} \\
            \mphubert-\feat \\
            \quad zero-shot & K-means (L\num{9})  & \num{63.24} & \num{69.78} & \num{54.71} & \num{68.11} \\
            \quad zero-shot & feat. freq. & \num{55.55} & \num{58.74} & \num{51.52} & \num{61.16} \\
            \quad + self-supervised FT & K-means (L\num{10})  & \num{67.34} & \num{76.05} & \num{55.26} & \num{66.72} \\
            \mphubert-\phone \\
            \quad zero-shot & K-means (proj) & \num{61.67} & \num{66.87} & \num{53.32} & \num{65.60} \\
            \quad zero-shot$^\diamond$ & one-hot (all phones) & \num{57.91} & \num{62.22} & \num{52.20} & \num{63.30} \\
            \quad + self-supervised FT & K-means (L\num{10})    & \num{67.74} & \num{76.46} & \underline{57.34} & \num{67.84} \\
            \bottomrule
            \multicolumn{2}{l}{$^\dagger$ Fetched from \citet{poli2025spidr}.} \\
        \end{tabular}%
    }
    \caption{\textbf{Spoken language modelling scores in English ($V = 256$).} Monolingual self-supervised learning baselines on top. $^\diamond$All phones seen in the data are kept ($V = 370$). Best scores in \textbf{bold}, second best \underline{underlined}.}
    \label{tab:slm-scores}
\end{table}

\paragraph{Results.}
\Cref{tab:slm-scores,tab:slm-scores-100} reveal a nuanced picture that does not always mirror the ABX probing results.
In zero-shot mode, \mphubert variants underperform the monolingual HuBERT-base and WavLM-base baselines on sWUGGY and sBLIMP, despite their superior phonetic discriminability in ABX.
This gap is consistent across both vocabulary sizes and suggests that the richer cross-lingual phonetic structure encoded by \mphubert does not translate straightforwardly into better lexical or syntactic priors for English-specific language modelling.
After self-supervised fine-tuning, however, both \mphubert variants recover competitive performance, approaching WavLM-base on sWUGGY (\qtyrange[ range-phrase = --, range-units = single]{67}{68}{\percent} vs. \qty{70}{\percent}) and matching or exceeding HuBERT-base across all three tasks.
This mirrors the pattern observed in ABX scores, where self-supervised fine-tuning substantially closes the gap with toplines.
The \mphubert-\phone variant with K-means fine-tuning achieves the strongest sLM performance among our models at $V=100$ (\Cref{tab:slm-scores-100}: \qty{67.65}{\percent} sWUGGY, \qty{76.48}{\percent} sBLIMP), consistent with its leading ABX scores in \Cref{tab:abx-scores}.

Taken together, these results suggest that while \mphubert's multilingual phonetic inductive biases do not directly boost English sLM performance in zero-shot mode, the representations become competitive after even minimal self-supervised adaptation, paralleling the conclusions drawn from ABX probing.
This conclusion is only tentative, because the base speech model was trained in English itself and is therefore difficult to disentangle the effect of the pre-training from our fine-tuning.
Further studies should start with a base universal model trained on languages other than English, and/or tested on a larger set of held-out languages.

\begin{table}[t]
    \centering
    \resizebox{0.47\textwidth}{!}{%
        \addtolength{\tabcolsep}{-0.2em}
        \begin{tabular}{lccccc}
            \toprule
            \multirow{2}{*}{Model} & \multirow{2}{*}{Units} & \multicolumn{2}{c}{sWUGGY} & \multirow{2}{*}{sBLIMP} & \multirow{2}{*}{tSC} \\
            \cmidrule(lr){3-4}
             & & all & in-vocab & & \\
            \midrule
            \mphubert-\feat \\
            \quad zero-shot & K-means (L\num{9})  & \num{64.63} & \num{71.85} & \num{56.15} & \num{68.00} \\
            \quad  zero-shot    & feat. freq. & \num{56.00} & \num{59.31} & \num{51.46} & \num{60.10} \\
            \quad + self-supervised FT & K-means (L\num{10})  & \underline{\num{67.08}} & \underline{\num{75.36}} & \num{55.89} & \num{67.74} \\
            \quad + self-supervised FT & one-hot (logits) & \num{66.83} & \num{75.34} & \underline{\num{56.24}} & \underline{\num{70.41}} \\
            \mphubert-\phone \\
            \quad zero-shot & K-means (proj) & \num{63.63} & \num{70.66} & \num{55.11} & \num{67.09} \\
            \quad zero-shot & phone freq. & \num{58.49} & \num{63.20} & \num{52.88} & \num{65.01} \\
            \quad + self-supervised FT & K-means (L\num{10}) & \textbf{\num[text-series-to-math]{67.65}} & \textbf{\num[text-series-to-math]{76.48}} & \textbf{\num[text-series-to-math]{58.83}} & \textbf{\num[text-series-to-math]{70.46}} \\
            \quad + self-supervised FT$^\diamond$ & one-hot (logits)  & \num{60.79} & \num{66.69} & \num{54.65} & \num{69.18} \\
            \bottomrule
        \end{tabular}%
    }
    \caption{\textbf{Spoken language modelling scores in English ($V = 100$).} $^\diamond$Only the phones seen during fine-tuning are used as vocabulary ($V = 87$). Best scores in \textbf{bold}, second best \underline{underlined}.}
    \label{tab:slm-scores-100}
\end{table}

\section{DiscoPhon benchmark}

DiscoPhon \citep{poli2026_discophon} is a benchmark designed to evaluate the ability of speech representation models to encode phonemic information on unlabelled data.
It consists of 6 development languages (German, Swahili, Tamil, Thai, Turkish and Ukrainian) and 6 test languages (Mandarin, English, Basque, French, Japanese and Wolof), selected to span a diverse range of phonological categories.
These language splits differ from those used in \S \ref{sec:few-shot-adapt}.
While \mphubert has not seen any of the development languages, it retains some English influence from the HuBERT backbone and some Japanese bias from the \qty{12.1}{\hour} of Japanese in the training data.

We compare \mphubert against the same probing baselines as in \S \ref{sec:few-shot-adapt}: MMS-1B, XEUS, mHuBERT-147, and HuBERT-base, using the best layers identified in \Cref{tab:abx-scores}.
We report results under three conditions:
\begin{enumerate*}[label=(\roman*)]
    \item many-to-one with \num{256} units, where each unit resulting from the clustering process is mapped to the most probable phoneme (\Cref{tab:discophon_100_m2o});
    \item one-to-one with ground-truth vocabulary size, where each phoneme is assigned a unique unit (\Cref{tab:discophon_100_o2o}); and
    \item one-to-one with fine-tuned models predicting the ground-truth vocabulary size, yielding a phoneme–unit bijection (\Cref{tab:discophon_0_o2o}). 
\end{enumerate*}
We evaluate using Phone Error Rate (PER), $R$-value, $F_1$, Phoneme-Normalised Mutual Information (PNMI), and triphone-based ABX discriminability on continuous representations.

\begin{table*}[t]
  \centering
  \resizebox{\textwidth}{!}{%
    \addtolength{\tabcolsep}{-0.25em}
    \begin{tabular}{lcccccccccc}
    \toprule
    & \multicolumn{5}{c}{dev languages} & \multicolumn{5}{c}{test languages} \\
    \cmidrule(lr){2-6} \cmidrule(lr){7-11}
    & PER $\downarrow$  & $R$-value $\uparrow$  & $F_1$ $\uparrow$  & PNMI $\uparrow$  & ABX c. $\downarrow$  & PER $\downarrow$  & $R$-value $\uparrow$  & $F_1$ $\uparrow$  & PNMI $\uparrow$  & ABX c. $\downarrow$ \\
    \midrule
    \quad\textbf{\textit{Zero-shot}} \\
    MMS-1B & \num{111.50} & \num{13.08} & \num{60.12} & \num{59.16} & \num{11.63} & \num{112.94} & \num{9.70} & \num{61.16} & \num{62.00} & \num{9.69} \\
    XEUS & \num{71.28} & \num{44.10} & \num{62.48} & \num{61.04} & \num{7.54} & \num{70.23} & \num{42.20} & \num{63.68} & \num{63.42} & \num{6.02} \\
    mHuBERT-147 & \num{70.16} & \num{47.87} & \num{66.51} & \num{63.96} & \num{9.16} & \num{71.21} & \num{43.88} & \num{68.01} & \num{67.14} & \num{7.33} \\
    HuBERT & \num{88.20} & \num{37.05} & \num{65.20} & \num{60.80} & \num{8.74} & \num{83.51} & \num{36.99} & \num{66.67} & \num{65.02} & \num{6.72} \\
    \cdashline{1-11}
    \mphubert-\feat & \textbf{\num[text-series-to-math]{56.52}} & \textbf{\num[text-series-to-math]{59.96}} & \underline{\num{72.75}} & \textbf{\num[text-series-to-math]{68.98}} & \underline{\num{7.18}} & \textbf{\num[text-series-to-math]{56.28}} & \textbf{\num[text-series-to-math]{57.34}} & \underline{\num{74.28}} & \textbf{\num[text-series-to-math]{72.10}} & \underline{\num{5.54}} \\
    \mphubert-\phone & \underline{\num{59.43}} & \underline{\num{58.09}} & \textbf{\num[text-series-to-math]{73.39}} & \underline{\num{68.59}} & \textbf{\num[text-series-to-math]{7.00}} & \underline{\num{60.84}} & \underline{\num{54.85}} & \textbf{\num[text-series-to-math]{74.92}} & \underline{\num{71.60}} & \textbf{\num[text-series-to-math]{5.20}} \\
    \addlinespace
    \quad\textbf{\textit{Self-supervised FT (10 h)}} \\
   HuBERT + K-means (L11) & \num{73.19} & \num{46.66} & \num{67.95} & \num{64.95} & \num{6.92} & \num{71.66} & \num{44.41} & \num{68.73} & \num{68.40} & \num{5.43} \\
         \cdashline{1-11}

    \mphubert-\feat \\
    \quad + K-means (L9) & \num{49.07} & \num{64.25} & \num{74.56} & \num{70.87} & \underline{\num{5.91}} & \num{47.43} & \num{62.84} & \num{76.06} & \num{74.04} & \num{4.60} \\
    \quad + K-means (feat) & \num{42.47} & \num{70.78} & \num{75.47} & \num{71.44} & \num{6.15} & \num{40.64} & \num{68.81} & \num{77.37} & \num{74.96} & \num{4.62} \\
    \quad + feat. freq. & \num{40.71} & \num{71.48} & \num{75.71} & \num{71.68} & \num{5.98} & \num{37.08} & \num{71.48} & \num{77.90} & \num{75.02} & \num{4.53} \\
    \quad + phone freq. & \num{38.31} & \num{74.20} & \num{77.61} & \num{72.44} & \num{6.04} & \num{36.83} & \num{72.33} & \num{79.66} & \num{75.58} & \num{4.66} \\
    \mphubert-\phone \\
    \quad + K-means (proj) & \num{50.98} & \num{62.98} & \num{74.84} & \num{70.93} & \num{6.10} & \num{50.33} & \num{61.01} & \num{76.54} & \num{73.90} & \num{4.53} \\
    \quad + K-means (phone) & \num{44.44} & \num{70.66} & \num{76.90} & \num{71.32} & \num{5.95} & \num{38.40} & \num{72.19} & \num{79.62} & \num{75.28} & \num{4.43} \\
    \quad + phone freq. & \textbf{\num[text-series-to-math]{34.09}} & \textbf{\num[text-series-to-math]{78.29}} & \textbf{\num[text-series-to-math]{80.06}} & \textbf{\num[text-series-to-math]{73.38}} & \num{5.93} & \underline{\num{32.70}} & \underline{\num{76.78}} & \underline{\num{82.05}} & \textbf{\num[text-series-to-math]{76.60}} & \textbf{\num[text-series-to-math]{4.41}} \\
    \quad + all phones & \underline{\num{35.22}} & \underline{\num{77.69}} & \underline{\num{79.92}} & \underline{\num{72.99}} & \textbf{\num[text-series-to-math]{5.89}} & \textbf{\num[text-series-to-math]{32.51}} & \textbf{\num[text-series-to-math]{77.09}} & \textbf{\num[text-series-to-math]{82.18}} & \underline{\num{76.48}} & \textbf{\num[text-series-to-math]{4.41}} \\
    \bottomrule
    \end{tabular}%
    }
    \caption{\textbf{DiscoPhon benchmark's many-to-one scores (\num{256} units).} Fine-tuned models predict \num{100} or \num{370} clusters (\mphubert-\phone + all phones MPR predicts all the seen phones). Units are mapped to the most probable phoneme for evaluation. Target layer as in \Cref{tab:abx-scores}. Results (in $\%$) averaged across dev. and test languages, resp. Triphone-based ABX averaged between within- and across-speaker conditions. K-means clustering: layer for pseudo-labels to be predicted between parentheses. Best scores in \textbf{bold}, second best \underline{underlined}.}
  \label{tab:discophon_100_m2o}
  \vspace{-0.5em}
\end{table*}

\begin{table*}[t]
  \centering
  \resizebox{\textwidth}{!}{%
    \addtolength{\tabcolsep}{-0.25em}
    \begin{tabular}{lcccccccc}
    \toprule
    & \multicolumn{4}{c}{dev languages} & \multicolumn{4}{c}{test languages} \\
    \cmidrule(lr){2-5} \cmidrule(lr){6-9}
    & PER $\downarrow$  & $R$-value $\uparrow$  & $F_1$ $\uparrow$  & PNMI $\uparrow$ & PER $\downarrow$  & $R$-value $\uparrow$  & $F_1$ $\uparrow$  & PNMI $\uparrow$ \\
    \midrule
    \quad\textbf{\textit{Zero-shot}} \\
    MMS-1B & \num{272.84} & \num{-106.13} & \num{42.91} & \num{42.64} & \num{270.69} & \num{-107.90} & \num{43.86} & \num{45.74} \\
    XEUS & \num{206.75} & \num{-53.59} & \num{44.46} & \num{43.40} & \num{215.05} & \num{-63.63} & \num{45.14} & \num{46.21} \\
    mHuBERT-147 & \num{220.10} & \num{-59.96} & \num{46.59} & \num{46.31} & \num{213.57} & \num{-58.99} & \num{48.09} & \num{49.17} \\
    HuBERT & \num{195.82} & \num{-40.63} & \num{50.33} & \num{47.68} & \num{199.22} & \num{-47.29} & \num{50.79} & \num{50.78} \\
                \cdashline{1-9}

    \mphubert-\feat & \underline{\num{134.26}} & \underline{\num{6.13}} & \underline{\num{58.50}} & \underline{\num{56.26}} & \underline{\num{146.20}} & \underline{\num{-8.15}} & \underline{\num{57.34}} & \underline{\num{58.75}} \\
    \mphubert-\phone & \textbf{\num[text-series-to-math]{123.24}} & \textbf{\num[text-series-to-math]{16.73}} & \textbf{\num[text-series-to-math]{62.58}} & \textbf{\num[text-series-to-math]{58.65}} & \textbf{\num[text-series-to-math]{128.29}} & \textbf{\num[text-series-to-math]{5.77}} & \textbf{\num[text-series-to-math]{61.85}} & \textbf{\num[text-series-to-math]{61.16}} \\
    \addlinespace
    \quad \textbf{\textit{Self-supervised FT (10 h)}} \\
    HuBERT + K-means (L11) & \num{175.23} & \num{-24.01} & \num{53.38} & \num{51.25} & \num{179.07} & \num{-33.25} & \num{53.11} & \num{53.59} \\
    \cdashline{1-9}
    \mphubert-\feat \\
    \quad + K-means (L9) & \num{129.80} & \num{8.84} & \num{59.44} & \num{57.31} & \num{144.22} & \num{-7.10} & \num{57.85} & \num{58.94} \\
    \quad + K-means (feat) & \num{88.39} & \num{44.93} & \num{66.51} & \num{61.25} & \num{102.86} & \num{29.27} & \num{65.28} & \num{63.67} \\
    \quad + feat. freq. & \num{86.80} & \num{47.02} & \num{67.45} & \num{61.42} & \num{100.97} & \num{32.16} & \num{65.82} & \num{63.70} \\
    \quad + phone freq. & \num{91.70} & \num{42.05} & \num{66.21} & \num{62.64} & \num{103.51} & \num{26.46} & \num{65.38} & \num{64.96} \\
    \mphubert-\phone \\
    \quad + K-means (proj) & \num{134.66} & \num{4.27} & \num{59.75} & \num{58.20} & \num{142.09} & \num{-7.41} & \num{58.63} & \num{60.06} \\
    \quad + K-means (phone) & \num{85.52} & \num{49.94} & \num{68.82} & \num{62.33} & \num{90.40} & \underline{\num{42.32}} & \num{69.27} & \num{65.16} \\
    \quad + phone freq. & \textbf{\num[text-series-to-math]{73.35}} & \textbf{\num[text-series-to-math]{58.24}} & \textbf{\num[text-series-to-math]{72.49}} & \textbf{\num[text-series-to-math]{64.84}} & \underline{\num{88.76}} & \num{39.20} & \underline{\num{69.57}} & \textbf{\num[text-series-to-math]{67.31}} \\
    \quad + all phones & \underline{\num{74.99}} & \underline{\num{57.78}} & \underline{\num{72.25}} & \underline{\num{64.12}} & \textbf{\num[text-series-to-math]{85.96}} & \textbf{\num[text-series-to-math]{44.09}} & \textbf{\num[text-series-to-math]{70.63}} & \underline{\num{66.89}} \\
    \bottomrule
    \end{tabular}%
    }
    \caption{\textbf{DiscoPhon benchmark's one-to-one scores (($|\mathcal{P}| + 1$) units).} Fine-tuned models predict \num{100} or \num{370} clusters (\mphubert-\phone + all phones MPR predicts all the seen phones). Each phoneme is mapped to a single unit. Target layer as in \Cref{tab:abx-scores}, best scores in \textbf{bold}, second best \underline{underlined}.}
  \label{tab:discophon_100_o2o}
  \vspace{-0.5em}
\end{table*}

\paragraph{Results.} 
The DiscoPhon results largely corroborate and extend the ABX findings from \S \ref{sec:few-shot-adapt}.
In zero-shot mode, both \mphubert variants substantially outperform all baselines across PER, $R$-value, $F_1$, and PNMI in the many-to-one condition (\Cref{tab:discophon_100_m2o}), with \mphubert-\feat and \mphubert-\phone reducing PER by at least \qty{15}{\percent} relative to the best multilingual baseline on development (mHuBERT-147) on test languages (XEUS).
This advantage is consistent with the phonetic invariance improvements observed in \Cref{tab:abx-scores}, and reinforces the conclusion that multilingual articulatory supervision yields genuinely better phoneme-level representations, not merely better triphone discriminability.

After self-supervised fine-tuning, \mphubert variants further strengthen their lead.
In the many-to-one condition (\Cref{tab:discophon_100_m2o}), \mphubert-\phone with phone frequency clustering achieves the best PER and $R$-value across both development and test languages (\qty{34.1}{\percent} PER, $R$-value \qty{78.3}{\percent} on dev; \qty{32.7}{\percent} PER, $R$-value \qty{76.8}{\percent} on test), paralleling its top ABX performance.
The feature frequency and phone frequency methods consistently outperform standard K-means, echoing the pattern from \S \ref{sec:few-shot-adapt-results}, where linguistically motivated clustering strategies showed advantages in longer temporal contexts.

The one-to-one condition (\Cref{tab:discophon_100_o2o,tab:discophon_0_o2o}) is considerably more demanding, as each phoneme must be assigned a dedicated unit.
Here, the ranking among methods is preserved, but absolute scores deteriorate sharply for all models, highlighting the difficulty of achieving a clean phoneme–unit bijection without gold supervision.
Notably, the K-means (phone) variant incurs a significant ABX cost in \Cref{tab:discophon_0_o2o}, suggesting that forcing a bijective mapping via K-means can hurt representational quality despite improving phoneme assignment metrics.
Nonetheless, \mphubert variants with phone- or feature-frequency-based fine-tuning remain the strongest systems (\Cref{tab:discophon_100_o2o}: \mphubert-\phone + phone freq. reaches \qty{73.4}{\percent} PER, \qty{58.2}{\percent} $R$-value on dev), and the ground-truth vocabulary fine-tuning condition (\Cref{tab:discophon_0_o2o}) shows that phone-frequency MPR yields the best trade-off between PER and $R$-value.

\begin{table*}[t]
  \centering
  \resizebox{\textwidth}{!}{%
    \addtolength{\tabcolsep}{-0.25em}
    \begin{tabular}{lcccccccccc}
    \toprule
    & \multicolumn{5}{c}{dev languages} & \multicolumn{5}{c}{test languages} \\
    \cmidrule(lr){2-6} \cmidrule(lr){7-11}
    & PER $\downarrow$  & $R$-value $\uparrow$  & $F_1$ $\uparrow$  & PNMI $\uparrow$  & ABX c. $\downarrow$  & PER $\downarrow$  & $R$-value $\uparrow$  & $F_1$ $\uparrow$  & PNMI $\uparrow$  & ABX c. $\downarrow$ \\
    \midrule
    HuBERT + K-means (L11) & \num{176.81} & \num{-24.97} & \num{52.84} & \num{48.96} & \num{9.18} & \num{182.38} & \num{-33.29} & \num{52.73} & \num{52.02} & \num{6.73} \\
                \cdashline{1-11}

    \mphubert-\feat \\
    \quad + K-means (L9) & \num{121.14} & \num{16.94} & \num{60.96} & \num{58.28} & \underline{\num{7.02}} & \num{133.60} & \num{1.74} & \num{59.37} & \num{60.45} & \underline{\num{5.26}} \\
    \quad + K-means (feat) & \num{104.38} & \num{40.92} & \num{62.63} & \num{47.65} & \num{11.16} & \num{110.65} & \num{31.75} & \num{64.03} & \num{53.30} & \num{8.08} \\
    \quad + feat. freq. & \num{86.26} & \num{53.96} & \num{65.72} & \num{46.44} & \num{8.71} & \num{89.50} & \num{48.24} & \num{67.36} & \num{51.43} & \num{6.66} \\
    \quad + phone freq. & \underline{\num{39.30}} & \underline{\num{81.98}} & \underline{\num{79.27}} & \num{59.88} & \num{8.29} & \underline{\num{40.22}} & \underline{\num{82.71}} & \underline{\num{82.69}} & \num{62.29} & \num{6.47} \\
    \mphubert-\phone \\
    \quad + K-means (proj) & \num{106.68} & \num{29.59} & \num{65.64} & \textbf{\num[text-series-to-math]{60.90}} & \textbf{\num[text-series-to-math]{6.66}} & \num{114.07} & \num{17.68} & \num{64.34} & \underline{\num{63.34}} & \textbf{\num[text-series-to-math]{4.76}} \\
    \quad + K-means (phone) & \num{101.66} & \num{52.07} & \num{64.56} & \num{48.87} & \num{17.67} & \num{103.08} & \num{46.18} & \num{67.10} & \num{55.38} & \num{11.41} \\
    \quad + phone freq. & \textbf{\num[text-series-to-math]{37.59}} & \textbf{\num[text-series-to-math]{82.93}} & \textbf{\num[text-series-to-math]{80.24}} & \underline{\num{60.83}} & \num{9.20} & \textbf{\num[text-series-to-math]{38.30}} & \textbf{\num[text-series-to-math]{83.82}} & \textbf{\num[text-series-to-math]{83.75}} & \textbf{\num[text-series-to-math]{63.73}} & \num{6.85} \\
    \bottomrule
  \end{tabular}%
  }
  \caption{\textbf{DiscoPhon benchmark's one-to-one scores ($|\mathcal{P}| + 1$ units).} Fine-tuned models predict the ground-truth vocabulary size, resulting in a phoneme-unit bijection. The target layer corresponds to the logits from the MPR task (on top of HuBERT's last layer). K-means clustering: layer for pseudo-labels to be predicted between parentheses. Best scores in \textbf{bold}, second best \underline{underlined}.}
  \label{tab:discophon_0_o2o}
  \vspace{-0.5em}
\end{table*}

\end{document}